\documentclass[runningheads]{llncs}
\usepackage{graphicx}
\usepackage{comment}
\usepackage{epsfig}
\usepackage{booktabs}
\usepackage{amsmath,amssymb,bm} 
\usepackage[font=small]{caption}
\usepackage{multirow}
\usepackage{subfig}
\usepackage{color}
\usepackage[dvipsnames]{xcolor}
\usepackage{hyperref}
\hypersetup{linkbordercolor=MidnightBlue}

\newcommand\blfootnote[1]{%
  \begingroup
  \renewcommand\thefootnote{}\footnote{#1}%
  \addtocounter{footnote}{-1}%
  \endgroup
}

\begin{document}
\pagestyle{headings}
\mainmatter
\def\ECCVSubNumber{2905}  

\title{View-Invariant Probabilistic Embedding for Human Pose} 


\titlerunning{View-Invariant Probabilistic Embedding for Human Pose}
%
\author{Jennifer~J.~Sun\index{Sun,Jennifer~J.}\inst{1,*} \and
Jiaping~Zhao\inst{2} \and
Liang-Chieh~Chen\inst{2} \and
Florian~Schroff\inst{2} \and
Hartwig~Adam\inst{2} \and
Ting~Liu\inst{2}}
\authorrunning{J.J.~Sun et al.}
%
\institute{California Institute of Technology \\
\email{jjsun@caltech.edu}
\and
Google Research\\
\email{\{jiapingz,lcchen,fschroff,hadam,liuti\}@google.com}}
\maketitle

\begin{abstract}

Depictions of similar human body configurations can vary with changing viewpoints. Using only 2D information, we would like to enable vision algorithms to recognize similarity in human body poses across multiple views. This ability is useful for analyzing body movements and human behaviors in images and videos. In this paper, we propose an approach for learning a compact view-invariant embedding space from 2D joint keypoints alone, without explicitly predicting 3D poses. Since 2D poses are projected from 3D space, they have an inherent ambiguity, which is difficult to represent through a deterministic mapping. Hence, we use probabilistic embeddings to model this input uncertainty. Experimental results show that our embedding model achieves higher accuracy when retrieving similar poses across different camera views, in comparison with 2D-to-3D pose lifting models. We also demonstrate the effectiveness of applying our embeddings to view-invariant action recognition and video alignment. Our code is available at \url{https://github.com/google-research/google-research/tree/master/poem}.

\vspace{-0.5cm}
\end{abstract}

\vspace{-0.2cm}
\section{Introduction}

\blfootnote{* This work was done during the author's internship at Google.}When we represent three dimensional (3D) human bodies in two dimensions (2D), the same human pose can appear different across camera views. There can be significant visual variations from a change in viewpoint due to changing relative depth of body parts and self-occlusions. Despite these variations, humans have the ability to recognize similar 3D human body poses in images and videos. This ability is useful for computer vision tasks where changing viewpoints should not change the labels of the task. We explore how we can embed 2D visual information of human poses to be consistent across camera views. We show that these embeddings are useful for tasks such as view-invariant pose retrieval, action recognition, and video alignment.

\begin{figure*}
    \centering
    \subfloat[View-Invariant Pose Embeddings (VIPE).\label{fig:goals:vipe}]{\includegraphics[width=0.4\textwidth]{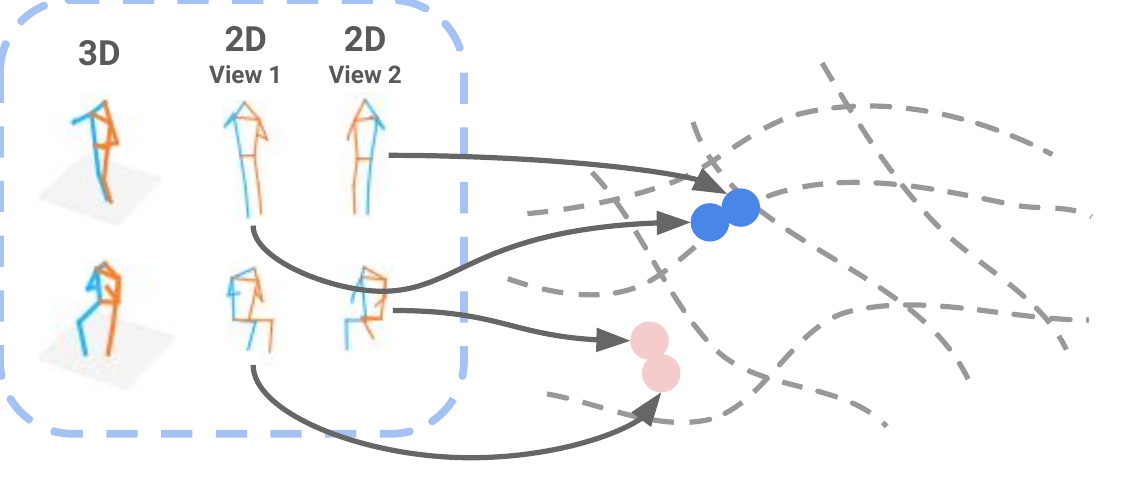}}\qquad
    \subfloat[Probabilistic View-Invariant Pose Embeddings (Pr-VIPE).\label{fig:goals:pr_vipe}]{\includegraphics[width=0.4\textwidth]{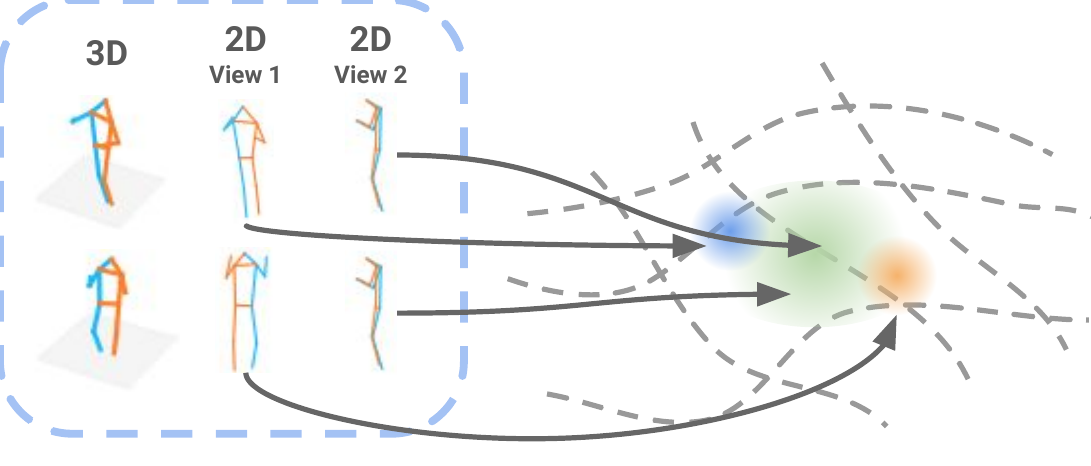}}
    \caption{We embed 2D poses such that our embeddings are (a) view-invariant (2D projections of similar 3D poses are embedded close) and (b) probabilistic (embeddings are distributions that cover different 3D poses projecting to the same input 2D pose).}
    \label{fig:goals}
    \vspace{-0.3cm}
\end{figure*}

Inspired by 2D-to-3D lifting models~\cite{martinez2017simple}, we learn view invariant embeddings directly from 2D pose keypoints. As illustrated in Fig.~\ref{fig:goals}, we explore whether view invariance of human bodies can be achieved from 2D poses alone, without predicting 3D pose. Typically, embedding models are trained from images using deep metric learning techniques~\cite{mori2015pose,ho2019pies,chu2019vehicle}. However, images with similar human poses can appear different because of changing viewpoints, subjects, backgrounds, clothing, etc. As a result, it can be difficult to understand errors in the embedding space from a specific factor of variation. Furthermore, multi-view image datasets for human poses are difficult to capture in the wild with 3D groundtruth annotations. In contrast, our method leverages existing 2D keypoint detectors: using 2D keypoints as inputs allows the embedding model to focus on learning view invariance. Our 2D keypoint embeddings can be trained using datasets in lab environments, while having the model generalize to in-the-wild data. Additionally, we can easily augment training data by synthesizing multi-view 2D poses from 3D poses through perspective projection. 

Another aspect we address is input uncertainty. The input to our embedding model is 2D human pose, which has an inherent ambiguity. Many valid 3D poses can project to the same or very similar 2D pose~\cite{akhter2015pose}. This input uncertainty is difficult to represent using deterministic mappings to the embedding space (point embeddings)~\cite{oh2018modeling,kendall2017uncertainties}. Our embedding space consists of probabilistic embeddings based on multivariate Gaussians, as shown in Fig.~\ref{fig:goals:pr_vipe}. We show that the learned variance from our method correlates with input 2D ambiguities. We call our approach Pr-VIPE for \textbf{Pr}obabilistic \textbf{V}iew-\textbf{I}nvariant \textbf{P}ose \textbf{E}mbeddings. The non-probabilistic, point embedding formulation will be referred to as VIPE. 

We show that our embedding is applicable to subsequent vision tasks such as pose retrieval~\cite{mori2015pose,jammalamadaka2012video}, video alignment~\cite{dwibedi2019temporal}, and action recognition~\cite{zhang2013actemes,iqbal2017pose}. One direct application is pose-based image retrieval. 
Our embedding enables users to search images by fine-grained pose, such as jumping with hands up, riding bike with one hand waving, and many other actions that are potentially difficult to pre-define. The importance of this application is further highlighted by works such as~\cite{mori2015pose,jammalamadaka2012video}.
Compared with using 3D keypoints with alignment for retrieval, our embedding enables efficient similarity comparisons in Euclidean space.

\textbf{Contributions} Our main contribution is the method for learning an embedding space where 2D pose embedding distances correspond to their similarities in absolute 3D pose space. We also develop a probabilistic formulation that captures 2D pose ambiguity.
We use cross-view pose retrieval to evaluate the view-invariant property: given a monocular pose image, we retrieve the same pose from different views without using camera parameters. Our results suggest 2D poses are sufficient to achieve view invariance without image context, and we do not have to predict 3D pose coordinates to achieve this. We also demonstrate the use of our embeddings for action recognition and video alignment.

\vspace{-0.2cm}
\section{Related Work}

\textbf{Metric Learning} We are working to understand similarity in human poses across views. Most works that aim to capture similarity between inputs generally apply techniques from metric learning. Objectives such as contrastive loss (based on pair matching)~\cite{bromley1994signature,hadsell2006dimensionality,oh2018modeling}  and triplet loss (based on tuple ranking)~\cite{wang2014learning,schroff2015facenet,wohlhart2015learning,hermans2017defense} are often used to push together/pull apart similar/dissimilar examples in embedding space. 
The number of possible training tuples increases exponentially with respect to the number of samples in the tuple, and not all combinations are equally informative.
To find informative training tuples, various mining strategies are proposed~\cite{schroff2015facenet,wu2017sampling,oh2016deep,hermans2017defense}. In particular, semi-hard triplet mining has been widely used~\cite{schroff2015facenet,wu2017sampling,parkhi2015deep}. This mining method finds negative examples that are fairly hard as to be informative but not too hard for the model. The hardness of a negative sample is based on its embedding distance to the anchor. Commonly, this distance is the Euclidean distance~\cite{wang2014learning,wohlhart2015learning,schroff2015facenet,hermans2017defense}, but any differentiable distance function could be applied~\cite{hermans2017defense}. \cite{huang2016local,iscen2018mining} show that alternative distance metrics also work for image and object retrieval. 

In our work, we learn a mapping from Euclidean embedding distance to a probabilistic similarity score. This probabilistic similarity captures closeness in 3D pose space from 2D poses. Our work is inspired by the mapping used in soft contrastive loss~\cite{oh2018modeling} for learning from an occluded N-digit MNIST dataset. 

Most of the papers discussed above involve deterministically mapping inputs to point embeddings. There are works that also map inputs to probabilistic embeddings. Probabilistic embeddings have been used to model specificity of word embeddings~\cite{vilnis2014word}, uncertainty in graph representations~\cite{bojchevski2017deep}, and input uncertainty due to occlusion~\cite{oh2018modeling}. We will apply probabilistic embeddings to address inherent ambiguities in 2D pose due to 3D-to-2D projection.

\textbf{Human Pose Estimation} 3D human poses in a global coordinate frame are view-invariant, since images across views are mapped to the same 3D pose. However, as mentioned by~\cite{martinez2017simple}, it is difficult to infer the 3D pose in an arbitrary global frame since any changes to the frame does not change the input data. 
Many approaches work with poses in the camera coordinate system~\cite{martinez2017simple,chen20173d,pavllo20193d,rayat2018exploiting,zhou2017towards,sun2018integral,rhodin2018unsupervised,tekin2017learning,chen2019unsupervised}, where the pose description changes based on viewpoint. While our work focuses on images with a single person, there are other works focusing on describing poses of multiple people~\cite{rhodin2019neural}.

Our approach is similar in setup to existing 3D lifting pose estimators~\cite{martinez2017simple,chen20173d,pavllo20193d,rayat2018exploiting,drover2018can} in terms of using 2D pose keypoints as input.
The difference is that lifting models are trained to regress to 3D pose keypoints, while our model is trained using metric learning and outputs an embedding distribution. Some recent works also use multi-view datasets to predict 3D poses in the global coordinate frame~\cite{qiu2019cross,kocabas2019self,iskakov2019learnable,rhodin2018learning,tome2018rethinking}. Our work differs from these methods with our goal (view-invariant embeddings), task (cross-view pose retrieval), and approach (metric learning). Another work on pose retrieval~\cite{mori2015pose} embeds images with similar 2D poses in the same view close together. Our method focuses on learning view invariance, and we also differ from~\cite{mori2015pose} in method (probabilistic embeddings).

\textbf{View Invariance and Object Retrieval} When we capture a 3D scene in 2D as images or videos, changing the viewpoint often does not change other properties of the scene. The ability to recognize visual similarities across viewpoints is helpful for a variety of vision tasks, such as motion analysis~\cite{ji2008visual,ji2009advances}, tracking~\cite{ong2006viewpoint}, vehicle and human re-identification~\cite{chu2019vehicle,zheng2019pose}, object classification and retrieval~\cite{lecun2004learning,hu2010learning,ho2019pies}, and action recognition~\cite{rao2001view,liu2018viewpoint,xia2012view,li2018unsupervised}.

Some of these works focus on metric learning for object retrieval. 
Their learned embedding spaces place different views of the same object class close together. Our work on human pose retrieval differs in a few ways. Our labels are continuous 3D poses, whereas in object recognition tasks, each embedding is associated with a discrete class label. Furthermore, we embed 2D poses, while these works embed images. Our approach allows us to investigate the impact of input 2D uncertainty with probabilistic embeddings and explore confidence measures to cross-view pose retrieval. We hope that our work provides a novel perspective on view invariance for human poses.

\begin{figure*}[t!]
  \centering
  \includegraphics[width=0.9\textwidth]{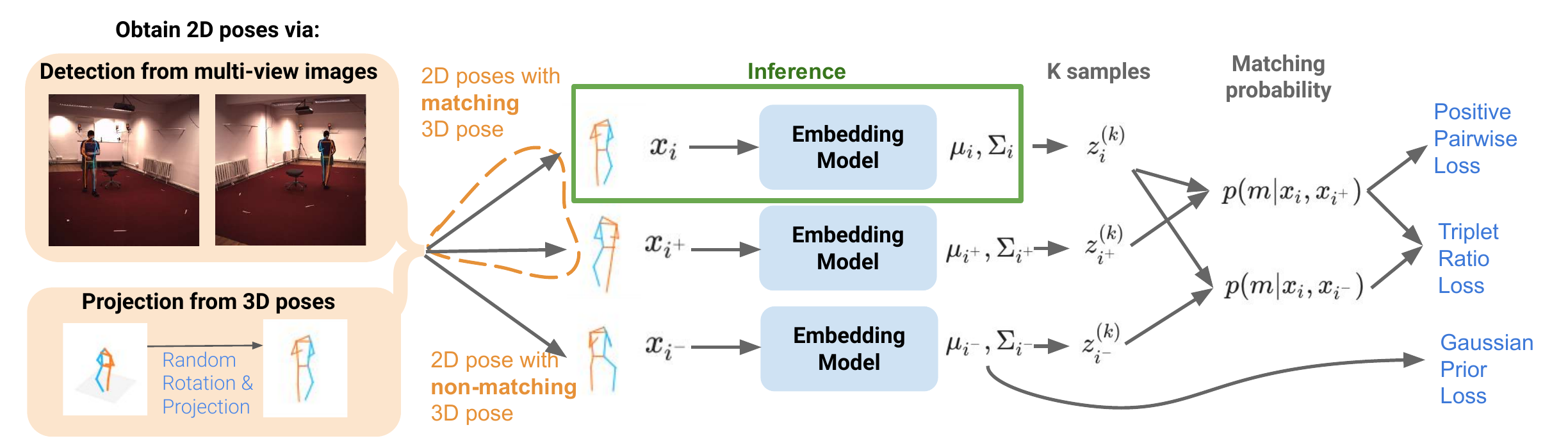}
  \caption{Overview of Pr-VIPE model training and inference. Our model takes keypoint input from a single 2D pose (detected from images and/or projected from 3D poses) and predicts embedding distributions. Three losses are applied during training.}
 \label{fig:2}
 \vspace{-0.3cm}
\end{figure*}

\vspace{-0.2cm}
\section{Our Approach}

The training and inference framework of Pr-VIPE is illustrated in Fig.~\ref{fig:2}. Our goal is to embed 2D poses such that distances in the embedding space correspond to similarities of their corresponding absolute 3D poses in Euclidean space. We achieve this view invariance property through our triplet ratio loss (Section~\ref{sec:triplet_ratio_loss}), which pushes together/pull apart 2D poses corresponding to similar/dissimilar 3D poses. The positive pairwise loss (Section~\ref{sec:positive_pairwise_loss}) is applied to increase the matching probability of similar poses. Finally, the Gaussian prior loss (Section~\ref{sec:probabilistic_embeddings}) helps regularize embedding magnitude and variance.

\vspace{-0.2cm}
\subsection{Matching Definition}\label{sec:matching_definition}

The 3D pose space is continuous, and two 3D poses can be trivially different without being identical. We define two 3D poses to be matching if they are visually similar regardless of viewpoint.
Given two sets of 3D keypoints $(\bm{y}_i,\bm{y}_j)$, we define a matching indicator function
\vspace{-0.2cm}
\begin{equation} \label{eq:12}
 \vspace{-0.1cm}
m_{ij} = \begin{cases}
    1, & \text{if } \text{NP-MPJPE}(\bm{y}_i, \bm{y}_j) \leqslant \kappa\\
    0,              & \text{otherwise,}
\end{cases}
\end{equation}
where $\kappa$ controls visual similarity between matching poses.
Here, we use mean per joint position error (MPJPE)~\cite{ionescu2013human3} between the two sets of 3D pose keypoints as a proxy to quantify their visual similarity. Before computing MPJPE, we normalize the 3D poses and apply Procrustes alignment between them. The reason is that we want our model to be view-invariant and to disregard rotation, translation, or scale differences between 3D poses. We refer to this normalized, Procrustes aligned MPJPE as \textbf{NP-MPJPE}.

\vspace{-0.2cm}
\subsection{Triplet Ratio Loss}\label{sec:triplet_ratio_loss}

The triplet ratio loss aims to embed 2D poses based on the matching indicator function (\ref{eq:12}). Let $n$ be the dimension of the input 2D pose keypoints $\bm{x}$, and $d$ be the dimension of the output embedding. We would like to learn a mapping $f: \mathbb{R}^n \rightarrow \mathbb{R}^d$, such that $D(\bm{z}_i, \bm{z}_j) < D(\bm{z}_i, \bm{z}_{j^{\prime}}),\forall m_{ij}>m_{ij^{\prime}}$, where $\bm{z} = f(\bm{x})$, and $D(\bm{z}_i, \bm{z}_j)$ is an embedding space distance measure.

For a pair of input 2D poses $(\bm{x}_i,\bm{x}_j)$, we define $p(m|\bm{x}_i,\bm{x}_j)$ to be the probability that their corresponding 3D poses $(\bm{y}_i,\bm{y}_j)$ match, that is, they are visually similar. While it is difficult to define this probability directly, we propose to assign its values by estimating $p(m|\bm{z}_i,\bm{z}_j)$ via metric learning. We know that if two 3D poses are identical, then $p(m|\bm{x}_i,\bm{x}_j)=1$, and if two 3D poses are sufficiently different, $p(m|\bm{x}_i,\bm{x}_j)$ should be small. 
For any given input triplet $(\bm{x}_i,\bm{x}_{i^+},\bm{x}_{i^-})$ with $m_{i,i^+}>m_{i,i^-}$, we want
\vspace{-0.2cm}
\begin{equation}
\frac{p(m|\bm{z}_i, \bm{z}_{i^+})}{p(m|\bm{z}_i,\bm{z}_{i^-})}\geqslant\beta,
\end{equation}
where $\beta>1$ represents the ratio of the matching probability of a similar 3D pose pair to that of a dissimilar pair. 
Applying negative logarithm to both sides, we have
\vspace{-0.2cm}
\begin{equation}
\left(-\log p(m|\bm{z}_i, \bm{z}_{i^+})\right) - \left(-\log p(m|\bm{z}_i, \bm{z}_{i^-})\right)\leqslant-\log\beta.
\end{equation}
Notice that the model can be trained to satisfy this with the triplet loss framework~\cite{schroff2015facenet}. Given batch size $N$, we define triplet ratio loss $\mathcal{L}_\text{ratio}$ as
\vspace{-0.2cm}
\begin{equation}\label{eq:ratio_loss}
\mathcal{L}_\text{ratio}=\sum_{i=1}^N\max(0,D_m(\bm{z}_i,\bm{z}_{i^+})-D_m(\bm{z}_i,\bm{z}_{i^-})+\alpha)),
\end{equation}
with distance kernel $D_m(\bm{z}_i,\bm{z}_j)=-\log p(m|\bm{z}_i,\bm{z}_j)$ and margin $\alpha=\log\beta$. To form a triplet $(\bm{x}_i,\bm{x}_{i^+},\bm{x}_{i^-})$, we set the anchor $\bm{x}_i$ and positive $\bm{x}_{i^+}$ to be projected from the same 3D pose and perform online semi-hard negative mining~\cite{schroff2015facenet} to find $\bm{x}_{i^-}$.

It remains for us to compute matching probability using our embeddings. To compute $p(m|\bm{z}_i, \bm{z}_j)$, we use the formulation proposed by~\cite{oh2018modeling}:
\begin{equation} \label{eq:6}
p(m|\bm{z}_i, \bm{z}_j) = \sigma(-a ||\bm{z}_i - \bm{z}_j||_2 + b),
\end{equation}
where $\sigma$ is a sigmoid function, and the trainable scalar parameters $a>0$ and $b\in\mathbb{R}$ calibrate embedding distances to probabilistic similarity. 

\vspace{-0.2cm}
\subsection{Positive Pairwise Loss}\label{sec:positive_pairwise_loss}

The positive pairs in our triplets have identical 3D poses. We would like them to have high matching probabilities, which can be encouraged by adding the positive pairwise loss
\vspace{-0.2cm}
\begin{equation} \label{eq:7}
\mathcal{L}_\text{positive} = \sum_{i=1}^N - \log p(m|\bm{z}_i, \bm{z}_{i^+}).
\end{equation}

The combination of $\mathcal{L}_\text{ratio}$ and $\mathcal{L}_\text{positive}$ can be applied to training point embedding models, which we refer to as VIPE in this paper.

\vspace{-0.2cm}
\subsection{Probabilistic Embeddings}\label{sec:probabilistic_embeddings}
In this section, we discuss the extension of VIPE to the probabilistic formulation Pr-VIPE. The inputs to our model, 2D pose keypoints, are inherently ambiguous, and there are many valid 3D poses projecting to similar 2D poses~\cite{akhter2015pose}. This input uncertainty can be difficult to model using point embeddings~\cite{kendall2017uncertainties,oh2018modeling}. We investigate representing this uncertainty using distributions in the embedding space by mapping 2D poses to probabilistic embeddings: $\bm{x} \rightarrow p(\bm{z}|\bm{x})$. Similar to~\cite{oh2018modeling}, we extend the input matching probability~(\ref{eq:6}) to using probabilistic embeddings as $p(m|\bm{x}_i,\bm{x}_j)=\int p(m|\bm{z}_i,\bm{z}_j)p(\bm{z}_i|\bm{x}_i)p(\bm{z}_j|\bm{x}_j)\textrm{d}\bm{z}_i\textrm{d}\bm{z}_j$, which can be approximated using Monte-Carlo sampling with $K$ samples drawn from each distribution as
\vspace{-0.2cm}
\begin{equation} \label{eq:9}
p(m|\bm{x}_i, \bm{x}_j) \approx \frac{1}{K^2} \sum_{k_1 = 1}^K \sum_{k_2 = 1}^K p(m|\bm{z}_i^{(k_1)}, \bm{z}_j^{(k_2)}).
\end{equation}

We model $p(\bm{z}|\bm{x})$ as a $d$-dimensional Gaussian with a diagonal covariance matrix. The model outputs mean $\mu(\bm{x})\in\mathbb{R}^d$ and covariance $\Sigma(\bm{x})\in\mathbb{R}^d$ with shared base network and different output layers. We use the reparameterization trick~\cite{kingma2013auto} during sampling.

In order to prevent variance from collapsing to zero and to regularize embedding mean magnitudes, we place a unit Gaussian prior on our embeddings with KL divergence by adding the Gaussian prior loss
\begin{equation} \label{eq:10}
\mathcal{L}_\text{prior} =\sum_{i=1}^N D_\text{KL}(\mathcal{N}(\mu(\bm{x}_i), \Sigma(\bm{x}_i))\:\|\: \mathcal{N}(\bm{0},\bm{I})).
\end{equation}

\textbf{Inference} At inference time, our model takes a single 2D pose (either from detection or projection) and outputs the mean and the variance of the embedding Gaussian distribution. 

\vspace{-0.2cm}
\subsection{Camera Augmentation}\label{sec:camera_augmentation}

Our triplets can be made of detected and/or projected 2D keypoints as shown in Fig.~\ref{fig:2}. When we train only with detected 2D keypoints, we are constrained to the camera views in training images. To reduce overfitting to these camera views, we perform camera augmentation by generating triplets using detected keypoints alongside projected 2D keypoints at random views.

To form triplets using multi-view image pairs, we use detected 2D keypoints from different views as anchor-positive pairs.
To use projected 2D keypoints, we perform two random rotations to a normalized input 3D pose to generate two 2D poses from different views for anchor/positive. Camera augmentation is then performed by using a mixture of detected and projected 2D keypoints. We find that training using camera augmentation can help our models learn to generalize better to unseen views (Section~\ref{sec:pose_retrieval_eval}).

\vspace{-0.2cm}
\subsection{Implementation Details}\label{sec:training_details}

We normalize 3D poses similar to~\cite{chen2019unsupervised}, and we perform instance normalization to 2D poses. The backbone network architecture for our model is based on~\cite{martinez2017simple}. We use $d=16$ as a good trade-off between embedding size and accuracy. To weigh different losses, we use $w_\text{ratio}=1$, $w_\text{positive}=0.005$, and $w_\text{prior}=0.001$. We choose $\beta=2$ for the triplet ratio loss margin and $K=20$ for the number of samples. The matching NP-MPJPE threshold is $\kappa=0.1$ for all training and evaluation. Our approach does not rely on a particular 2D keypoint detector, and we use PersonLab \cite{papandreou2018personlab} for our experiments. For random rotation in camera augmentation, we uniformly sample azimuth angle between $\pm180^{\circ}$, elevation between $\pm30^{\circ}$, and roll between $\pm30^{\circ}$. Our implementation is in TensorFlow, and all the models are trained with CPUs. More details and ablation studies on hyperparamters are provided in the supplementary materials.

\vspace{-0.2cm}
\section{Experiments}

We demonstrate the performance of our model through pose retrieval across different camera views (Section~\ref{sec:retrieval_subsection}). We further show our embeddings can be directly applied to downstream tasks, such as action recognition (Section~\ref{sec:action_recognition}) and video alignment (Section~\ref{sec:sequence_alignment}), without any additional training.

\vspace{-0.2cm}
\subsection{Datasets}\label{sec:datasets}

For all the experiments in this paper, we only train on a subset of the Human3.6M~\cite{ionescu2013human3} dataset. For pose retrieval experiments, we validate on the Human3.6M hold-out set and test on another dataset (MPI-INF-3DHP~\cite{mehta2017monocular}), which is unseen during training and free from parameter tuning.
We also present qualitative results on MPII Human Pose~\cite{andriluka20142d}, for which 3D groundtruth is not available. Additionally, we directly use our embeddings for action recognition and sequence alignment on Penn Action~\cite{zhang2013actemes}.

\textbf{Human3.6M (H3.6M)} H3.6M is a large human pose dataset recorded from $4$ chest level cameras with 3D pose groundtruth. We follow the standard protocol~\cite{martinez2017simple}: train on Subject 1, 5, 6, 7, and 8, and hold out Subject 9 and 11 for validation. For evaluation, we remove near-duplicate 3D poses within $0.02$ NP-MPJPE, resulting in a total of $10910$ evaluation frames per camera. This process is camera-consistent, meaning if a frame is selected under one camera, it is selected under all cameras, so that the perfect retrieval result is possible. 

\textbf{MPI-INF-3DHP (3DHP)} 3DHP is a more recent human pose dataset that contains $14$ diverse camera views and scenarios, covering more pose variations than H3.6M~\cite{mehta2017monocular}. We use $11$ cameras from this dataset and exclude the $3$ cameras with overhead views. 
Similar to H3.6M, we remove near-duplicate 3D poses, resulting in $6824$ frames per camera. We use all 8 subjects from the train split of 3DHP. \textbf{This dataset is only used for testing.}

\textbf{MPII Human Pose (2DHP)} This dataset is commonly used in 2D pose estimation, containing $25$K images from YouTube videos. Since groundtruth 3D poses are not available, we show qualitative results on this dataset.

\textbf{Penn Action} This dataset contains $2326$ trimmed videos for $15$ pose-based actions from different views. We follow the standard protocol~\cite{nie2015joint} for our action classification and video alignment experiments.

\vspace{-0.2cm}
\subsection{View-Invariant Pose Retrieval}\label{sec:retrieval_subsection}

Given multi-view human pose datasets, we query using detected 2D keypoints from one camera view and find the nearest neighbors in the embedding space from a different camera view. We iterate through all camera pairs in the dataset as query and index. Results averaged across all cameras pairs are reported.

\vspace{-0.2cm}
\subsubsection{Evaluation Procedure}

We report Hit@$k$ with $k=1$, $10$, and $20$ on pose retrievals, which is the percentage of top-$k$ retrieved poses that have at least one accurate retrieval. A retrieval is considered accurate if the 3D groundtruth from the retrieved pose satisfies the matching function~(\ref{eq:12}) with $\kappa=0.1$.

\textbf{Baseline Approaches} We compare Pr-VIPE with 2D-to-3D lifting models~\cite{martinez2017simple} and $L2$-VIPE. $L2$-VIPE outputs $L2$-normalized point embeddings, and is trained with the squared $L2$ distance kernel, similar to~\cite{schroff2015facenet}.

For fair comparison, we use the same backbone network architecture for all the models. Notably, this architecture~\cite{martinez2017simple} has been tuned for lifting tasks on H3.6M. Since the estimated 3D poses in camera coordinates are not view-invariant, we apply normalization and Procrustes alignment to align the estimated 3D poses between index and query for retrieval. In comparison, our embeddings do not require any alignment or other post-processing during retrieval. 

For Pr-VIPE, we retrieve poses using nearest neighbors in the embedding space with respect to the sampled matching probability~(\ref{eq:9}), which we refer to as retrival confidence. We present the results on the VIPE models with and without camera augmentation. 
We applied similar camera augmentation to the lifting model, but did not see improvement in performance. 
We also show the results of pose retrieval using aligned 2D keypoints only. The poor performance of using input 2D keypoints for retrieval from different views confirms the fact that models must learn view invariance from inputs for this task.

We also compare with the image-based EpipolarPose model~\cite{kocabas2019self}. Please refer to the supplementary materials for the experiment details and results.

\vspace{-0.2cm}
\subsubsection{Quantitative Results}\label{sec:pose_retrieval_eval}

\begin{table*}[!t]
  \centering
\caption{Comparison of cross-view pose retrieval results Hit@$k$ ($\%$) on H3.6M and 3DHP with chest-level cameras and all cameras. $*$ indicates that normalization and Procrustes alignment are performed on query-index pairs.} \label{tab:3dhp}  
   \begin{tabular}{c | c c c | c c c | c c c} 
   \toprule[0.2em]
   \multicolumn{1}{c|}{Dataset} & \multicolumn{3}{c|}{H3.6M} & \multicolumn{3}{c|}{3DHP (Chest)} & \multicolumn{3}{c}{3DHP (All)} \\
  $k$ & $1$ & $10$ & $20$ & $1$ & $10$ & $20$ & $1$ & $10$ & $20$  \\
   \toprule[0.2em]
   2D keypoints* & $28.7$ & $47.1$ & $50.9$ & $5.20$ &  $14.0$ & $17.2$ & $9.80$ & $21.6$ & $25.5$ \\
   3D lifting* & $69.0$ & $89.7$ & $92.7$ & $24.9$ & $54.4$ & $62.4$ & $24.6$  & $53.2$ & $61.3$  \\
   $L2$-VIPE & $73.5$  & $94.2$ & $96.6$ & $23.8$ & $56.7$ & $66.5$ & $18.7$  & $46.3$ & $55.7$ \\
   $L2$-VIPE (w/ aug.) & $70.4$ & $91.8$ & $94.5$ & $24.9$  & $55.4$ & $63.6$ & $23.7 $  & $53.0$ & $61.4$  \\
   Pr-VIPE & $\bm{76.2}$ &  $\bm{95.6}$ & $\bm{97.7}$ & $25.4$ & $59.3$ & $69.3$ & $19.9$ & $49.1$ & $58.8$ \\
   Pr-VIPE (w/ aug.) & $73.7$ & $93.9$ & $96.3$ & $\bm{28.3}$ & $\bm{62.3}$ & $\bm{71.4}$ & $\bm{26.4}$  & $\bm{58.6}$ & $\bm{67.9}$ \\
   \bottomrule[0.1em]
\end{tabular}
\vspace{-0.3cm}
\end{table*}

From Table~\ref{tab:3dhp}, we see that Pr-VIPE (with augmentation) outperforms all the baselines for H3.6M and 3DHP. The H3.6M results shown are on the hold-out set, and 3DHP is unseen during training, with more diverse poses and views.
When we use all the cameras from 3DHP, we evaluate the generalization ability of models to new poses and new views. When we evaluate using only the 5 chest-level cameras from 3DHP, where the views are more similar to the training set in H3.6M, we mainly evaluate for generalization to new poses.
When we evaluate using only the $5$ chest-level cameras from 3DHP, the views are more similar to H3.6M, and generalization to new poses becomes more important. 
Our model is robust to the choice of $\beta$ and the number of samples $K$ (analysis in supplementary materials).

Table~\ref{tab:3dhp} shows that Pr-VIPE without camera augmentation is able to perform better than the baselines for H3.6M and 3DHP (chest-level cameras). This shows that Pr-VIPE is able to generalize as well as other baseline methods to new poses. However, for 3DHP (all cameras), the performance for Pr-VIPE without augmentation is worse compared with chest-level cameras. This observation indicates that when trained on chest-level cameras only, Pr-VIPE does not generalize as well to new views. The same results can be observed for $L2$-VIPE between chest-level and all cameras. In contrast, the 3D lifting models are able to generalize better to new views with the help of additional Procrustes alignment, which requires expensive SVD computation for every index-query pair.

We further apply camera augmentation to training the Pr-VIPE and the $L2$-VIPE model. Note that this step does not require camera parameters or additional groundtruth. The results in Table~\ref{tab:3dhp} on Pr-VIPE show that the augmentation improves performance for 3DHP (all cameras) by $6\%$ to $9\%$. This step also increases chest-level camera accuracy slightly. For $L2$-VIPE, we can observe a similar increase on all views. Camera augmentation reduces accuracy on H3.6M for both models. This is likely because augmentation reduces overfitting to the training camera views.
By performing camera augmentation, Pr-VIPE is able to generalize better to new poses and new views.

\def\figsize{0.13}
\def\fighspace{0mm}
\def\fighspacer{+1.5mm}
\begin{figure*}[t!]
\centering
\begin{tabular}{cccccc}
\centering
\scriptsize{$C=0.960$}\hspace{\fighspace} & \scriptsize{$E=0.001$}\hspace{\fighspacer} & \scriptsize{$C=0.993$}\hspace{\fighspace} & \scriptsize{$E=0.098$}\hspace{\fighspacer} & \scriptsize{$C=0.983$}\hspace{\fighspace} & \scriptsize{$E=0.172$}\hspace{\fighspacer} \\
\includegraphics[width=\figsize\textwidth]{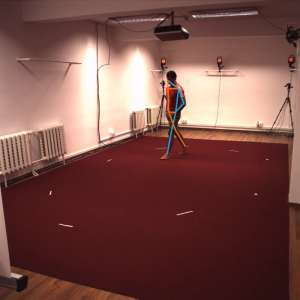}\hspace{\fighspace} & \includegraphics[width=\figsize\textwidth]{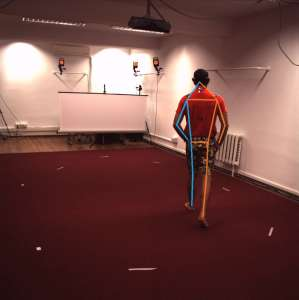}\hspace{\fighspacer}  & \includegraphics[width=\figsize\textwidth]{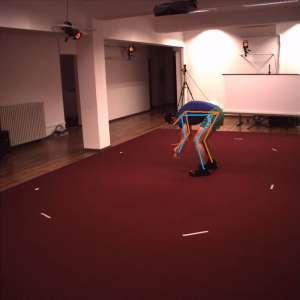}\hspace{\fighspace} & \includegraphics[width=\figsize\textwidth]{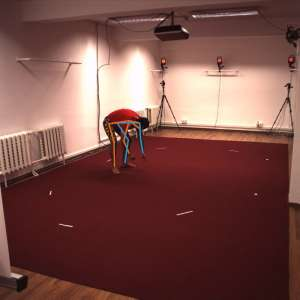}\hspace{\fighspacer} &
\includegraphics[width=\figsize\textwidth]{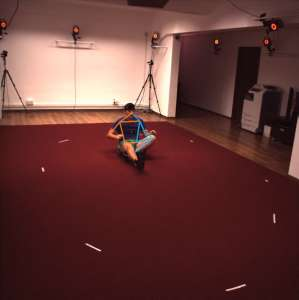}\hspace{\fighspace} & \includegraphics[width=\figsize\textwidth]{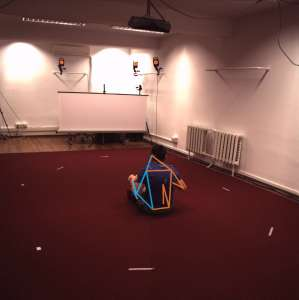}\hspace{\fighspacer} \\

\scriptsize{$C=0.651$}\hspace{\fighspace} & \scriptsize{$E=0.055$}\hspace{\fighspacer} & \scriptsize{$C=0.774$}\hspace{\fighspace} & \scriptsize{$E=0.082$}\hspace{\fighspacer} & \scriptsize{$C=0.426$}\hspace{\fighspace} & \scriptsize{$E=0.230$}\hspace{\fighspacer} \\
\includegraphics[width=\figsize\textwidth]{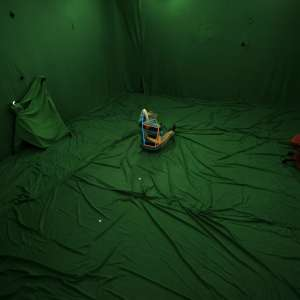}\hspace{\fighspace} & \includegraphics[width=\figsize\textwidth]{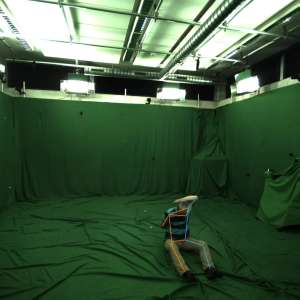}\hspace{\fighspacer}  & \includegraphics[width=\figsize\textwidth]{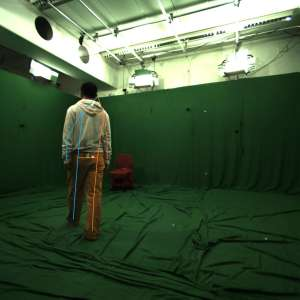}\hspace{\fighspace} & \includegraphics[width=\figsize\textwidth]{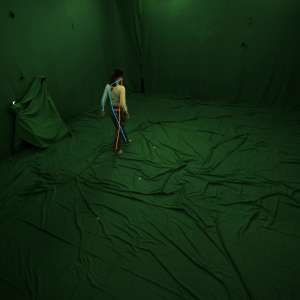}\hspace{\fighspacer} &
\includegraphics[width=\figsize\textwidth]{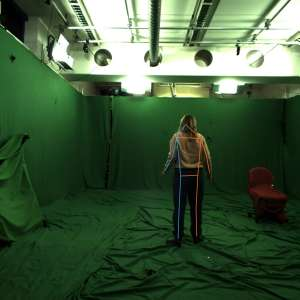}\hspace{\fighspace} & \includegraphics[width=\figsize\textwidth]{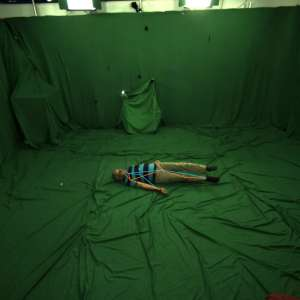}\hspace{\fighspacer} \\

\scriptsize{$C=0.629$}\hspace{\fighspace} & \scriptsize{$E=0.151$}\hspace{\fighspacer} & \scriptsize{$C=0.969$}\hspace{\fighspace} & \scriptsize{$E=0.034$}\hspace{\fighspacer} & \scriptsize{$C=0.808$}\hspace{\fighspace} & \scriptsize{$E=0.471$}\hspace{\fighspacer} \\
\includegraphics[width=\figsize\textwidth]{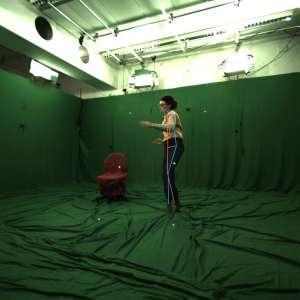}\hspace{\fighspace} & \includegraphics[width=\figsize\textwidth]{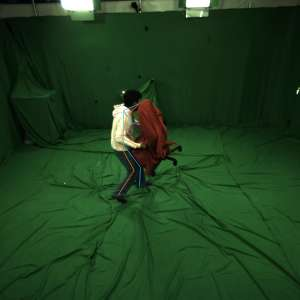}\hspace{\fighspacer}  & \includegraphics[width=\figsize\textwidth]{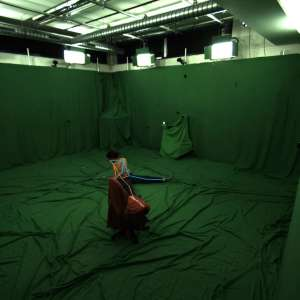}\hspace{\fighspace} & \includegraphics[width=\figsize\textwidth]{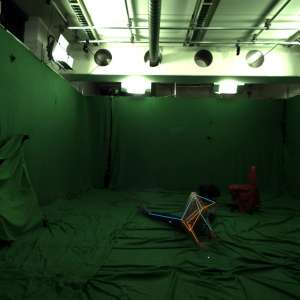}\hspace{\fighspacer} &
\includegraphics[width=\figsize\textwidth]{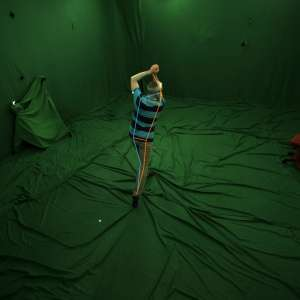}\hspace{\fighspace} & \includegraphics[width=\figsize\textwidth]{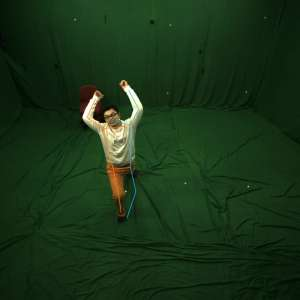}\hspace{\fighspacer} \\

\scriptsize{$C=0.963$}\hspace{\fighspace} & \hspace{\fighspacer} & \scriptsize{$C=0.599$}\hspace{\fighspace} & \hspace{\fighspacer} & \scriptsize{$C=0.914$}\hspace{\fighspace} & \hspace{\fighspacer} \\
\includegraphics[width=\figsize\textwidth]{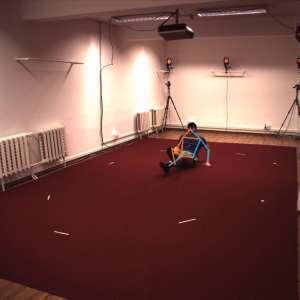}\hspace{\fighspace} & \includegraphics[width=\figsize\textwidth]{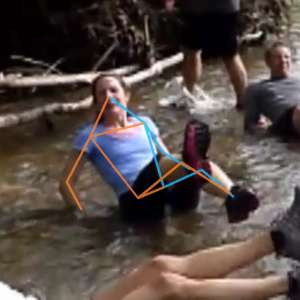}\hspace{\fighspacer}  & \includegraphics[width=\figsize\textwidth]{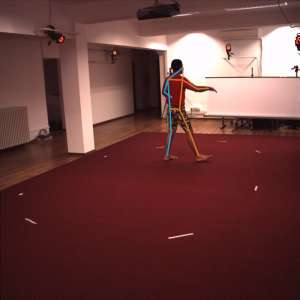}\hspace{\fighspace} & \includegraphics[width=\figsize\textwidth]{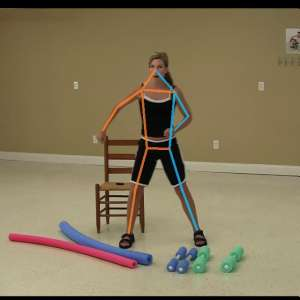}\hspace{\fighspacer} &
\includegraphics[width=\figsize\textwidth]{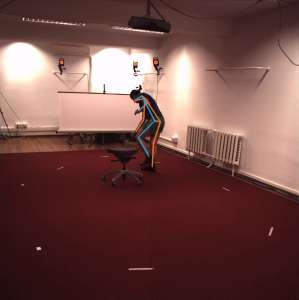}\hspace{\fighspace} & \includegraphics[width=\figsize\textwidth]{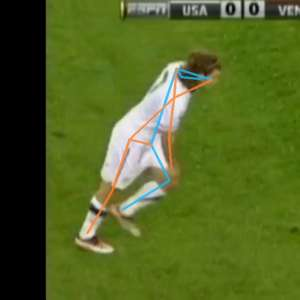}\hspace{\fighspacer} \\

\scriptsize{$C=0.957$}\hspace{\fighspace} & \hspace{\fighspacer} & \scriptsize{$C=0.987$}\hspace{\fighspace} & \hspace{\fighspacer} & \scriptsize{$C=0.877$}\hspace{\fighspace} & \hspace{\fighspacer} \\
\includegraphics[width=\figsize\textwidth]{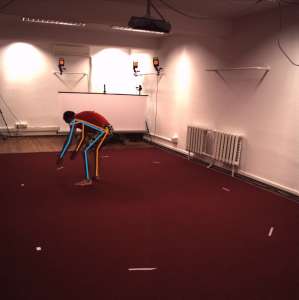}\hspace{\fighspace} & \includegraphics[width=\figsize\textwidth]{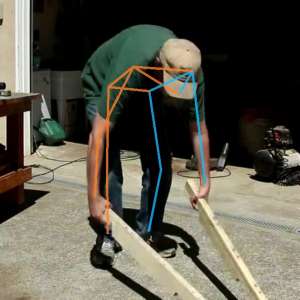}\hspace{\fighspacer}  & \includegraphics[width=\figsize\textwidth]{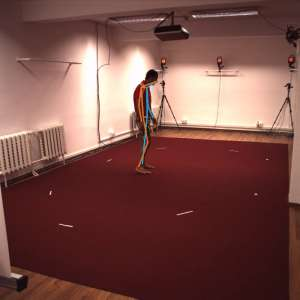}\hspace{\fighspace} & \includegraphics[width=\figsize\textwidth]{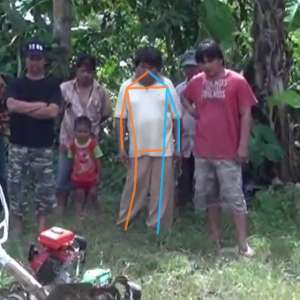}\hspace{\fighspacer} &
\includegraphics[width=\figsize\textwidth]{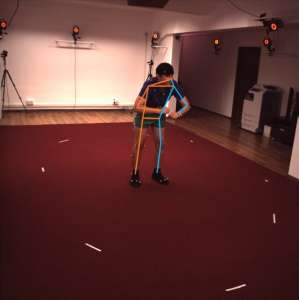}\hspace{\fighspace} & \includegraphics[width=\figsize\textwidth]{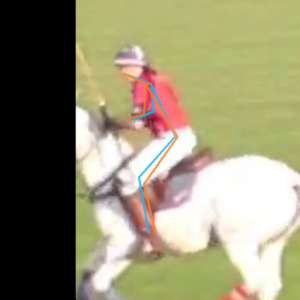}\hspace{\fighspacer} \\

\end{tabular}
\caption{Visualization of pose retrieval results. The first row is from H3.6M; the second and the third row are from 3DHP; the last two rows are using queries from H3.6M to retrieve from 2DHP. On each row, we show the query pose on the left for each image pair and the top-$1$ retrieval using the Pr-VIPE model (w/ aug.) on the right. We display retrieval confidences (``$C$'') and top-$1$ NP-MPJPEs (``$E$'', if 3D pose groundtruth is available).}
\label{fig:retrieval}
\vspace{-0.3cm}
\end{figure*}

\vspace{-0.2cm}
\subsubsection{Qualitative Results} 

Fig.~\ref{fig:retrieval} shows qualitative retrieval results using Pr-VIPE. 
As shown in the first row, the retrieval confidence of the model is generally high for H3.6M. This indicates that the retrieved poses are close to their queries in the embedding space.
Errors in 2D keypoint detection can lead to retrieval errors as shown by the rightmost pair. In the second and third rows, the retrieval confidence is lower for 3DHP. 
This is likely because there are new poses and views unseen during training, which has the nearest neighbor slightly further away in the embedding space. 
We see that the model can generalize to new views as the images are taken at different camera elevations from H3.6M. Interestingly, the rightmost pair on row 2 shows that the model can retrieve poses with large differences in roll angle, which is not present in the training set. The rightmost pair on row 3 shows an example of a large NP-MPJPE error due to mis-detection of the left leg in the index pose.

We show qualitative results using queries from the H3.6M hold-out set to retrieve from 2DHP in the last two rows of Fig.~\ref{fig:retrieval}. The results on these in-the-wild images indicate that as long as the 2D keypoint detector works reliably, our model is able to retrieve poses across views and subjects. 
More qualitative results are provided in the supplementary materials.

\vspace{-0.2cm}
\subsection{Downstream Tasks}\label{sec:tasks}

We show that our pose embedding can be directly applied to pose-based downstream tasks using simple algorithms. We compare the performance of Pr-VIPE (\textbf{only trained on H3.6M, with no additional training}) on the Penn Action dataset against other approaches specifically trained for each task on the target dataset. In all the following experiments in this section, we compute our Pr-VIPE embeddings on single video frames and use the negative logarithm of the matching probability (\ref{eq:9}) as the distance between two frames. Then we apply temporal averaging within an atrous kernel of size $7$ and rate $3$ around the two center frames and use this averaged distance as the frame matching distance. Given the matching distance, we use standard dynamic time warping (DTW) algorithm to align two action sequences by minimizing the sum of frame matching distances. We further use the averaged frame matching distance from the alignment as the distance between two video sequences.

\vspace{-0.2cm}
\subsubsection{Action Recognition}\label{sec:action_recognition}

We evaluate our embeddings for action recognition using nearest neighbor search with the sequence distance described above. Provided person bounding boxes in each frame, we estimate 2D pose keypoints using~\cite{papandreou2017towards}. On Penn Action, we use the standard train/test split~\cite{nie2015joint}. Using all the testing videos as queries, we conduct two experiments: (1) we use all training videos as index to evaluate overall performance and compare with state-of-the-art methods, and (2) we use training videos only under one view as index and evaluate the effectiveness of our embeddings in terms of view-invariance. For this second experiment, actions with zero or only one sample under the index view are ignored, and accuracy is averaged over different views.

\begin{table}[t]
    \centering
    \begin{minipage}[b]{0.45\textwidth}
    \centering
    \caption{Comparison of action recognition results on Penn Action.}\label{tab:action_recognition}
    \scalebox{0.8}{
    \begin{tabular}{c|ccc|c}
        \toprule[0.2em]
        \multirow{2}{*}{Methods} & \multicolumn{3}{c|}{Input} & \multirow{2}{*}{Accuracy ($\%$)}\\
        & RGB & Flow & Pose & \\
        \toprule[0.2em]
        Nie~\textit{et al.}~\cite{nie2015joint} & \checkmark & & \checkmark & $85.5$ \\
        Iqbal~\textit{et al.}~\cite{iqbal2017pose} & & & \checkmark & $79.0$\\
        Cao~\textit{et al.}~\cite{cao2017body} & & \checkmark & \checkmark & $95.3$ \\
        & \checkmark & \checkmark & & $98.1$ \\
        Du~\textit{et al.}~\cite{du2017rpan} & \checkmark & \checkmark & \checkmark & $97.4$ \\
        Liu~\textit{et al.}~\cite{liu2018recognizing} & \checkmark & & \checkmark & $91.4$ \\
        Luvizon~\textit{et al.}~\cite{luvizon2019multi} & \checkmark & & \checkmark & $98.7$ \\
        \textbf{Ours} & & & \checkmark & $97.5$\\
        \bottomrule[0.1em]
        Ours (1-view index) & & & \checkmark & $92.1$\\
        \bottomrule[0.1em]
    \end{tabular}
    }
    \end{minipage}\qquad\quad
    \begin{minipage}[b]{0.45\textwidth}
    \centering
    \caption{Comparison of video alignment results on Penn Action.} \label{tab:seq_align_res}
    \scalebox{0.8}{    
      \begin{tabular}{l | c  }
   \toprule[0.2em]
  Methods &  Kendall's Tau \\
   \toprule[0.2em]
   SaL \cite{misra2016shuffle} & $0.6336$\\
   TCN \cite{sermanet2018time} & $0.7353$\\
   TCC \cite{dwibedi2019temporal} & $0.7328$\\
   TCC + SaL \cite{dwibedi2019temporal} & $0.7286$\\
   TCC + TCN \cite{dwibedi2019temporal} & $0.7672$\\
   \textbf{Ours} & $0.7476$\\
    \bottomrule[0.1em] 
    Ours (same-view only) & $0.7521$\\
    Ours (different-view only) & $0.7607$\\
      \bottomrule[0.1em]
\end{tabular}
}
    \end{minipage}
\end{table}

From Table~\ref{tab:action_recognition} we can see that without any training on the target domain or using image context information, our embeddings can achieve highly competitive results on pose-based action classification, outperforming the existing best baseline that only uses pose input and even some other methods that rely on image context or optical flow. As shown in the last row in Table~\ref{tab:action_recognition}, our embeddings can be used to classify actions from different views using index samples from only one single view with relatively high accuracy, which further demonstrates the advantages of our view-invariant embeddings.

\vspace{-0.2cm}
\subsubsection{Video Alignment}\label{sec:sequence_alignment}

Our embeddings can be used to align human action videos from different views using DTW algorithm as described earlier in Section~\ref{sec:tasks}. We measure the alignment quality of our embeddings quantitatively using Kendall's Tau~\cite{dwibedi2019temporal}, which reflects how well an embedding model can be applied to align unseen sequences if we use nearest neighbor in the embedding space to match frames for video pairs. A value of $1$ corresponds to perfect alignment. We also test the view-invariant properties of our embeddings by evaluating Kendall's Tau on aligning videos pairs from the same view, and aligning pairs with different views.

\begin{figure*}[t!]
  \centering
  \includegraphics[width=\linewidth]{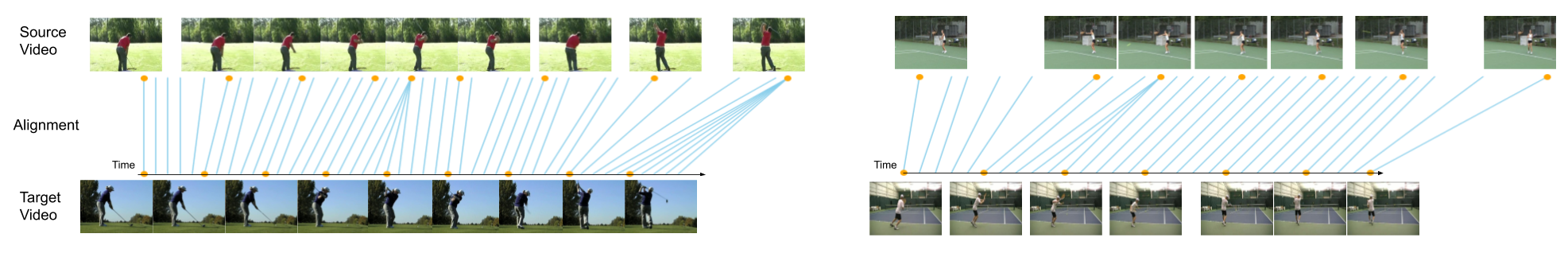}
  \caption{Video alignment results using Pr-VIPE. The orange dots correspond to the visualized frames, and the blue line segments illustrate the frame alignment.}
  \label{fig:video_sync}
  \vspace{-0.5cm}
\end{figure*}

In Table~\ref{tab:seq_align_res}, we compare our results with other video embedding baselines that are trained for the alignment task on Penn Action, from which we observe that Pr-VIPE performs better than all the method that use a single type of loss. While Pr-VIPE is slightly worse than the combined TCC+TCN loss, our embeddings are able to achieve this without being explicitly trained for this task or taking advantage of image context. In the last two rows of Table~\ref{tab:seq_align_res}, we show the results from evaluating video pairs only from the same or different views. We can see that our embedding achieves consistently high performance regardless of whether the aligned video pair is from the same or different views, which demonstrate its view-invariant property. 
In Fig.~\ref{fig:video_sync}, we show action video synchronization results from different views using Pr-VIPE. We provide more synchronized videos for all actions in the supplementary materials.

\vspace{-0.2cm}
\subsection{Ablation Study}\label{sec:ablation_studies}

\textbf{Point vs. Probabilistic Embeddings} We compare VIPE point embedding formulation with Pr-VIPE. When trained on detected keypoints, the Hit@$1$ for VIPE and Pr-VIPE are $75.4\%$ and $76.2\%$ on H3.6M, and $19.7\%$ and $20.0\%$ on 3DHP, respectively. When we add camera augmentation, the Hit@$1$ for VIPE and Pr-VIPE are $73.8\%$ and $73.7\%$ on H3.6M, and $26.1\%$ and $26.5\%$ on 3DHP, respectively.
Despite the similar retrieval accuracies, Pr-VIPE is generally more accurate and, more importantly, has additional desirable properties in that the variance can model 2D input ambiguity as to be discussed next.

A 2D pose is ambiguous if there are similar 2D poses that can be projected from very different poses in 3D. To measure this, we compute the average 2D NP-MPJPE between a 2D pose and its top-$10$ nearest neighbors in terms of 2D NP-MPJPE. To ensure the 3D poses are different, we sample $1200$ poses from H3.6M hold-out set with a minimum gap of $0.1$ 3D NP-MPJPE. If a 2D pose has small 2D NP-MPJPE to its neighbors, it means there are many similar 2D poses corresponding to different 3D poses and so the 2D pose is ambiguous.

\begin{figure*}[t!]
  \vspace{-0.4cm}
  \centering
  \subfloat[\label{fig:variance:pose}]{\includegraphics[width=0.6\textwidth]{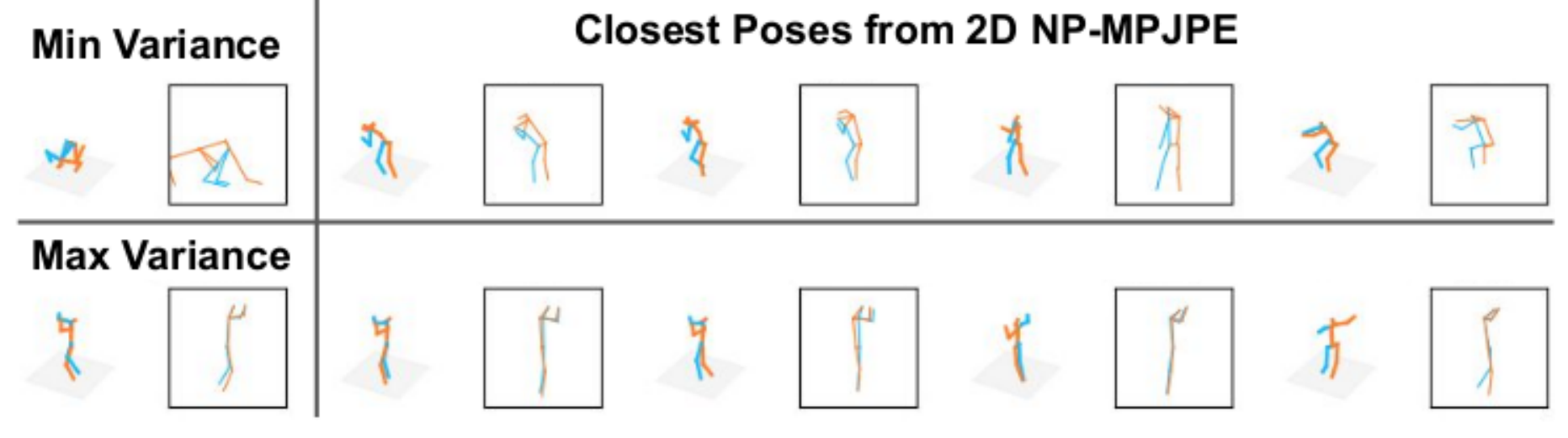}}\quad
  \subfloat[\label{fig:variance:plot}]{\includegraphics[width=0.35\textwidth]{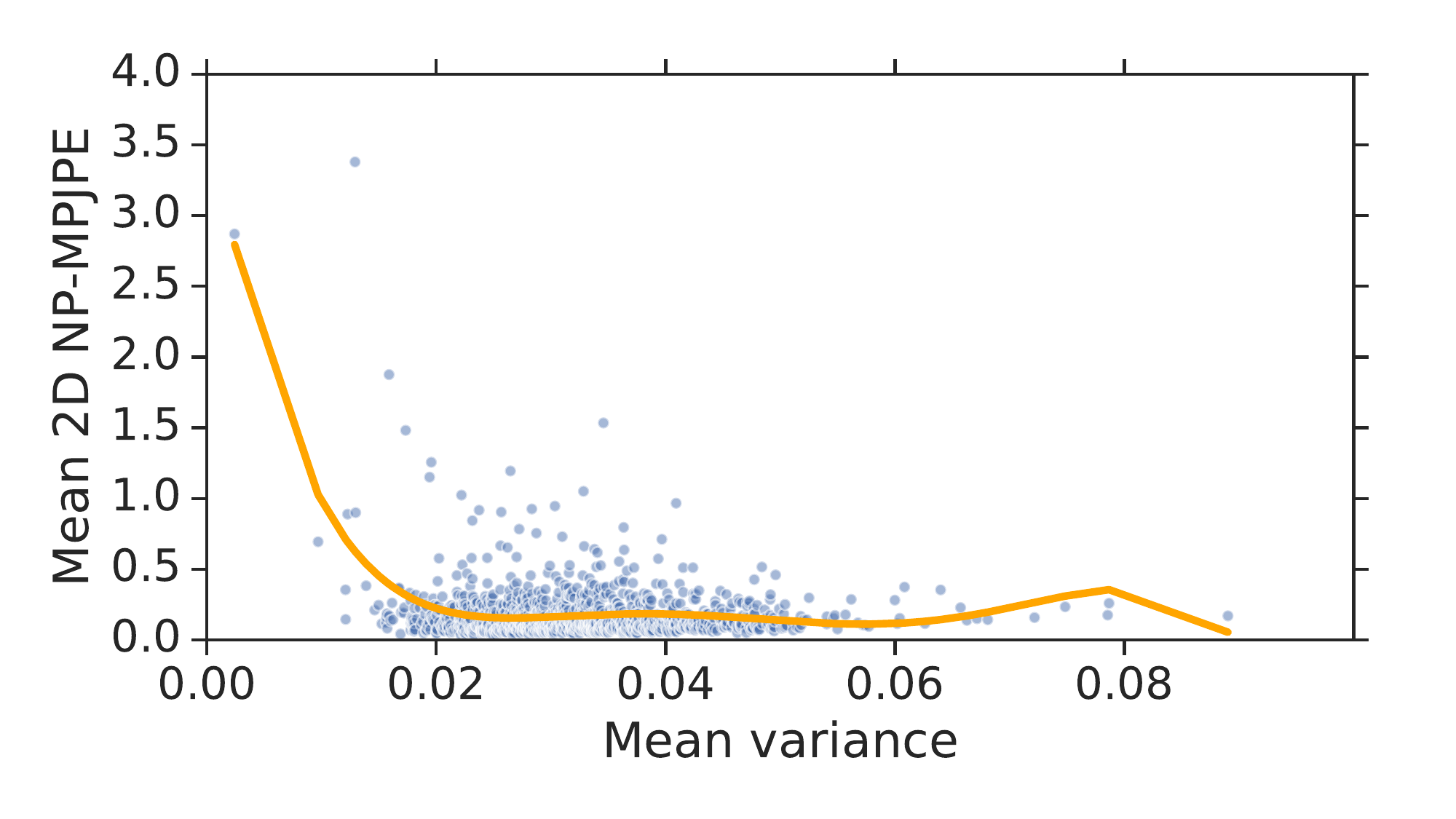}}\\
  \subfloat[\label{fig:emb_acc:dim}]{\includegraphics[width=0.35\textwidth]{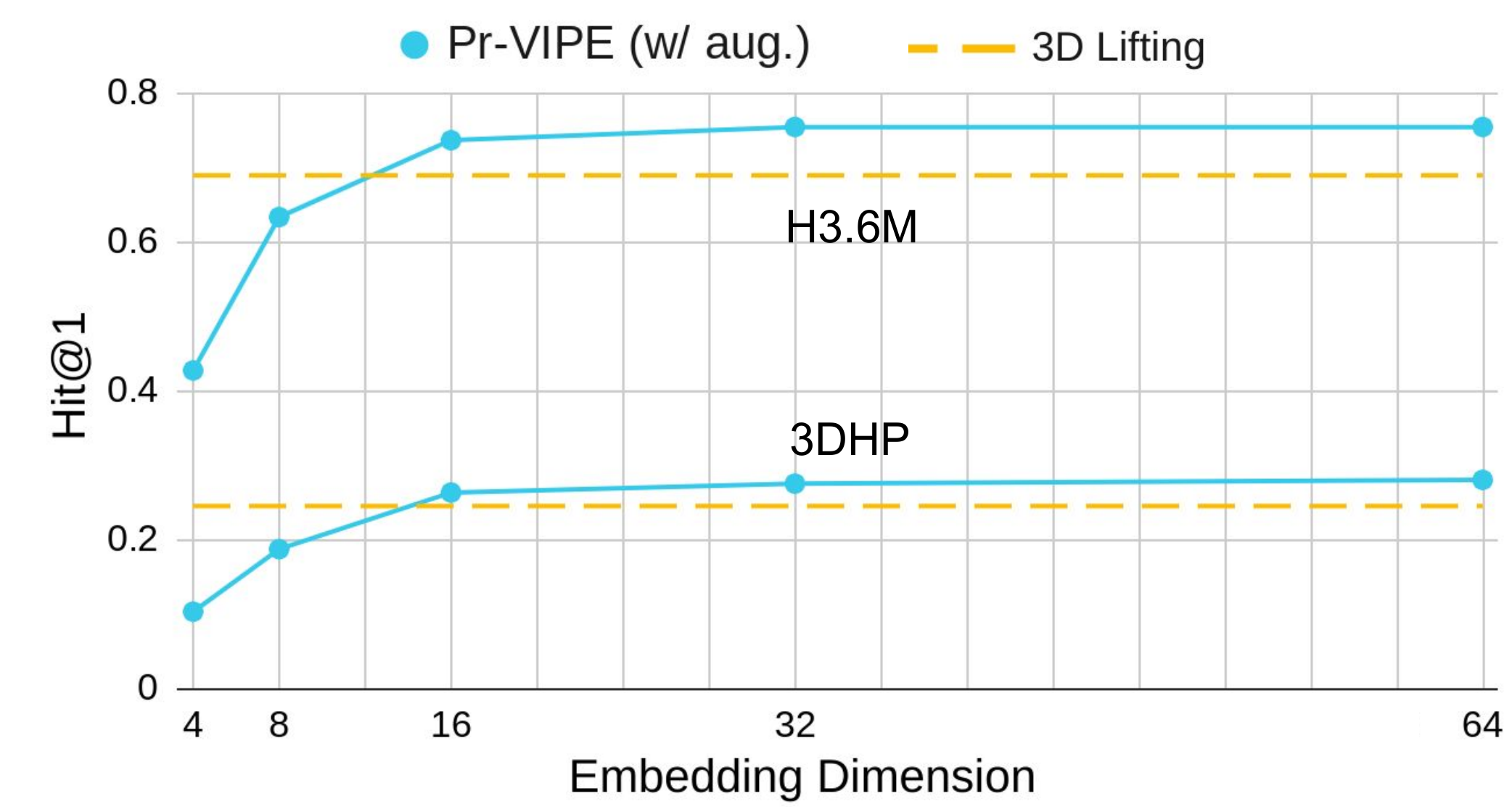}}\quad
  \subfloat[\label{fig:emb_acc:conf}]{\includegraphics[width=0.35\textwidth]{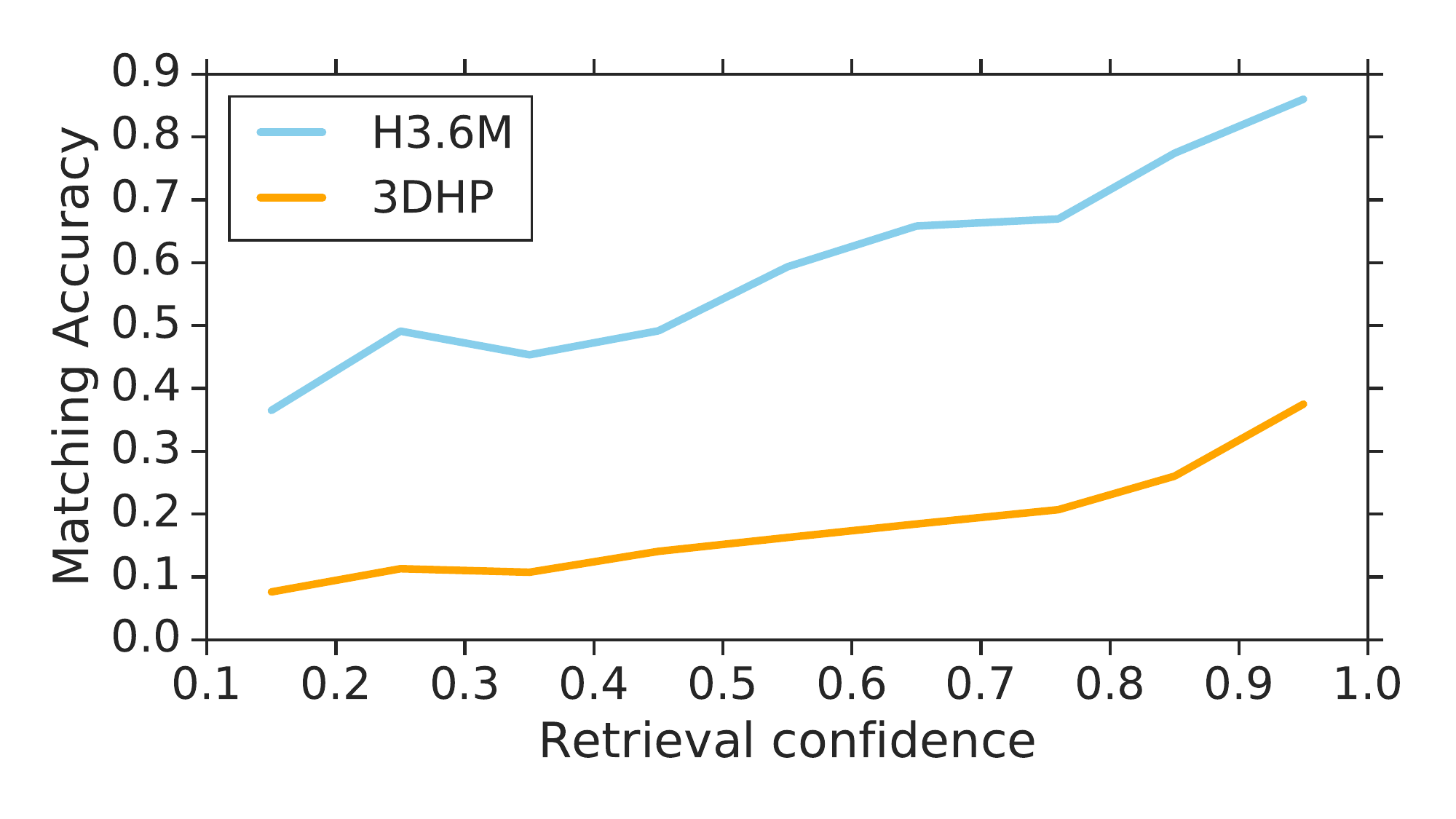}}
  \caption{Ablation study: (a) Top retrievals by 2D NP-MPJPE from the H3.6M hold-out subset for queries with largest and smallest variance. 2D poses are shown in the boxes. (b) Relationship between embedding variance and 2D NP-MPJPE to top-$10$ nearest 2D pose neighbors from the H3.6M hold-out subset. The orange curve represents the best fitting 5th degree polynomial. (c) Comparison of Hit@$1$ with different embedding dimensions. The 3D lifting baseline predictss $39$ dimensions. (d) Relationship between retrieval confidence and matching accuracy.}
  \vspace{-0.5cm}
  \label{fig:ablation_study}
\end{figure*}

Fig.~\ref{fig:variance:pose} shows that the 2D pose with the largest variance is ambiguous as it has similar 2D poses in H3.6M with different 3D poses. In contrast, we see that the closest 2D poses corresponding to the smallest variance pose on the first row of Fig.~\ref{fig:variance:pose} are clearly different. Fig.~\ref{fig:variance:plot} further shows that as the average variance increases, the 2D NP-MPJPE between similar poses generally decreases, which means that 2D poses with larger variances are more ambiguous.

\textbf{Embedding Dimensions} Fig.~\ref{fig:emb_acc:dim} demonstrates the effect of embedding dimensions on H3.6M and 3DHP. The lifting model lifts $13$ 2D keypoints to 3D, and therefore has a constant output dimension of $39$. We see that Pr-VIPE (with augmentation) is able to achieve a higher accuracy than lifting at $16$ dimensions. Additionally, we can increase the number of embedding dimensions to $32$, which increases accuracy of Pr-VIPE from $73.7\%$ to $75.5\%$.

\textbf{Retrieval Confidence} In order to validate the retrieval confidence values, we randomly sample $100$ queries along with their top-$5$ retrievals (using Pr-VIPE retrieval confidence) from each query-index camera pair. This procedure forms $6000$ query-retrieval sample pairs for H3.6M ($4$ views, $12$ camera pairs) and $55000$ for 3DHP ($11$ views, $110$ camera pairs), which we bin by their retrieval confidences. Fig.~\ref{fig:emb_acc:conf} shows the matching accuracy for each confidence bin. We can see that the accuracy positively correlates with the confidence values, which suggest our retrieval confidence is a valid indicator to model performance.

\textbf{What if 2D keypoint detectors were perfect? } We repeat our pose retrieval experiments using groundtruth 2D keypoints to simulate a perfect 2D keypoint detector on H3.6M and 3DHP. All experiments use the $4$ views from H3.6M for training following the standard protocol. For the baseline lifting model in camera frame, we achieve $89.9\%$ Hit@$1$ on H3.6M, $48.2\%$ on 3DHP (all), and $48.8\%$ on 3DHP (chest). For Pr-VIPE, we achieve $97.5\%$ Hit@$1$ on H3.6M, $44.3\%$ on 3DHP (all), and $66.4\%$ on 3DHP (chest). These results follow the same trend as using detected keypoints inputs in Table~\ref{tab:3dhp}. Comparing the results with using detected keypoints, the large improvement in performance using groundtruth keypoints suggests that a considerable fraction of error in our model is due to imperfect 2D keypoint detections. Please refer to the supplementary materials for more ablation studies and embedding space visualization.

\vspace{-0.2cm}
\section{Conclusion}

We introduce Pr-VIPE, an approach to learning probabilistic view-invariant embeddings from 2D pose keypoints.
By working with 2D keypoints, we can use camera augmentation to improve model generalization to unseen views.
We also demonstrate that our probabilistic embedding learns to capture input ambiguity.
Pr-VIPE has a simple architecture and can be potentially applied to object and hand poses. For cross-view pose retrieval, 3D pose estimation models require expensive rigid alignment between query-index pair, while our embeddings can be applied to compare similarities in simple Euclidean space. In addition, we demonstrated the effectiveness of our embeddings on downstream tasks for action recognition and video alignment. 
Our embedding focuses on a single person, and for future work, we will investigate extending it to multiple people and robust models that can handle missing keypoints from input.

\vspace{-0.2cm}
\section{Acknowledgment}

We thank Yuxiao Wang, Debidatta Dwibedi, and Liangzhe Yuan from Google Research, Long Zhao from Rutgers University, and Xiao Zhang from University of Chicago for helpful discussions. We appreciate the support of Pietro Perona, Yisong Yue, and the Computational Vision Lab at Caltech for making this collaboration possible. The author Jennifer J.\ Sun is supported by NSERC (funding number PGSD3-532647-2019) and Caltech.

\newpage
\appendix
\renewcommand\thefigure{\thesection\arabic{figure}}
\renewcommand{\thetable}{\thesection\arabic{table}}



\newcommand{\customizedparagraph}{\textbf}


\title{View-Invariant Probabilistic Embedding for Human Pose} 
\subtitle{Supplementary Materials}


\titlerunning{View-Invariant Probabilistic Embedding for Human Pose}
%
\author{Jennifer~J.~Sun\index{Sun,Jennifer~J.}\inst{1} \and
Jiaping~Zhao\inst{2} \and
Liang-Chieh~Chen\inst{2} \and
Florian~Schroff\inst{2} \and
Hartwig~Adam\inst{2} \and
Ting~Liu\inst{2}}

\authorrunning{J.J.~Sun et al.}
\institute{California Institute of Technology \\
\email{jjsun@caltech.edu}
\and
Google Research\\
\email{\{jiapingz,lcchen,fschroff,hadam,liuti\}@google.com}}
\maketitle

In this document, we cover the details of the implementation and experiments for our work. We also provide additional ablation studies and analysis. Specifically, we have:
\begin{itemize}
    \item[\textbullet] \hyperref[sec:3d_visual_similarity]{\textbf{\textcolor{MidnightBlue}{Section A}}} describes how we decide the \textbf{NP-MPJPE threshold} based on its effect on visual pose similarity.
    \item[\textbullet] \hyperref[sec:implementation_details]{\textbf{\textcolor{MidnightBlue}{Section B}}} provides additional \textbf{implementation details} on model training, keypoint definition and normalization, downstream task experiment setup, etc.
    \item[\textbullet] \hyperref[sec:ablation]{\textbf{\textcolor{MidnightBlue}{Section C}}} provides additional \textbf{ablation studies}, including the effect of key hyperparameters, ordered embedding variance visualizations, and embedding space visualization.
    \item[\textbullet] \hyperref[sec:additional_comp]{\textbf{\textcolor{MidnightBlue}{Section D}}} provides additional \textbf{quantitative pose retrieval} result comparisons with image-based EpipolarPose model~\cite{kocabas2019self} for view-invariant pose retrieval.
    \item[\textbullet] \hyperref[sec:pose_retrieval_qres]{\textbf{\textcolor{MidnightBlue}{Section E}}} provides additional \textbf{qualitative pose retrieval} results.
    \item[\textbullet] \hyperref[sec:video_alignment_qres]{\textbf{\textcolor{MidnightBlue}{Section F}}} describes the \textbf{qualitative video alignment} experiment. Please refer to {\scriptsize{\url{https://drive.google.com/open?id=1kTc_UT0Eq0H2ZBgfEoh8qEJMFBouC-Wv}}} for the video synchronization results.
\end{itemize}

\section{Visualization of 3D Visual Similarity}\label{sec:3d_visual_similarity}

The 3D pose space is continuous, and we use the NP-MPJPE as a proxy to quantify visual similarity between pose pairs. Fig.~\ref{fig:supp_similarity} shows pairs of 3D pose keypoints with their corresponding NP-MPJPE, where each row depicts a different NP-MPJPE range. This plot demonstrates the effect of choosing different $\kappa$, which controls matching threshold between 3D poses. If we choose $\kappa = 0.05$, then only the first row in Fig.~\ref{fig:supp_similarity} would be considered matching, and the rest of the rows are non-matching. Our current value of $\kappa = 0.10$ corresponds to using the first two rows as matching pairs and the rest of the rows as non-matching ones. By loosening $\kappa$, poses with greater differences will be considered as matching, as shown by different rows in Fig.~\ref{fig:supp_similarity}. We note that pairs in rows 3 and 4 shows significant visual differences compared with the first two rows. We further investigate the effects of different $\kappa$ during training and evaluation in Section~\ref{sec:ablation}.

\def\figsize{0.24}
\def\fighspace{1mm}
\def\fighspacer{-3mm}
\begin{figure*}[!t]
\centering
\begin{tabular}{cccccc}
\centering

\scriptsize{NP-MPJPE: $0.011$}\hspace{\fighspace} & \scriptsize{NP-MPJPE: $0.026$}\hspace{\fighspace} & \scriptsize{NP-MPJPE: $0.033$} & \scriptsize{NP-MPJPE: $0.049$}\hspace{\fighspace} \\

\includegraphics[width=\figsize\textwidth]{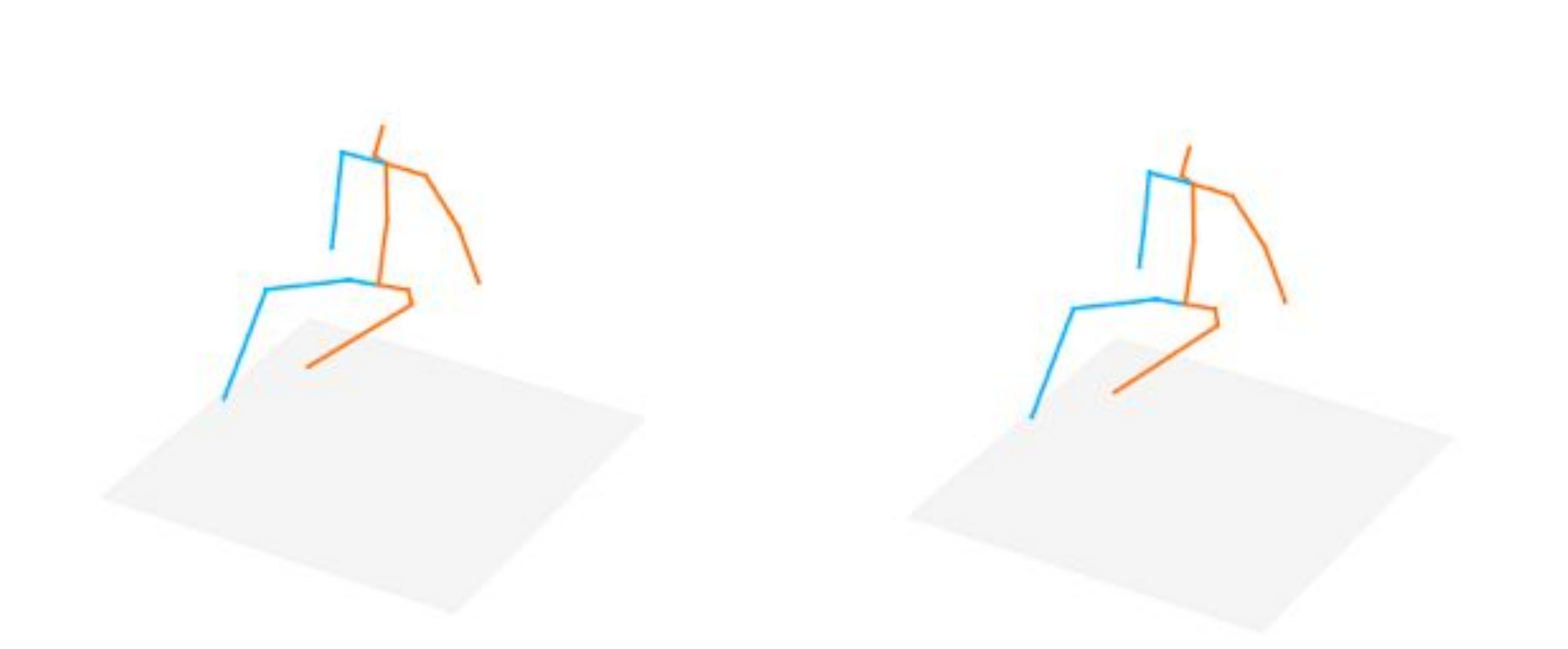}\hspace{\fighspace} & \includegraphics[width=\figsize\textwidth]{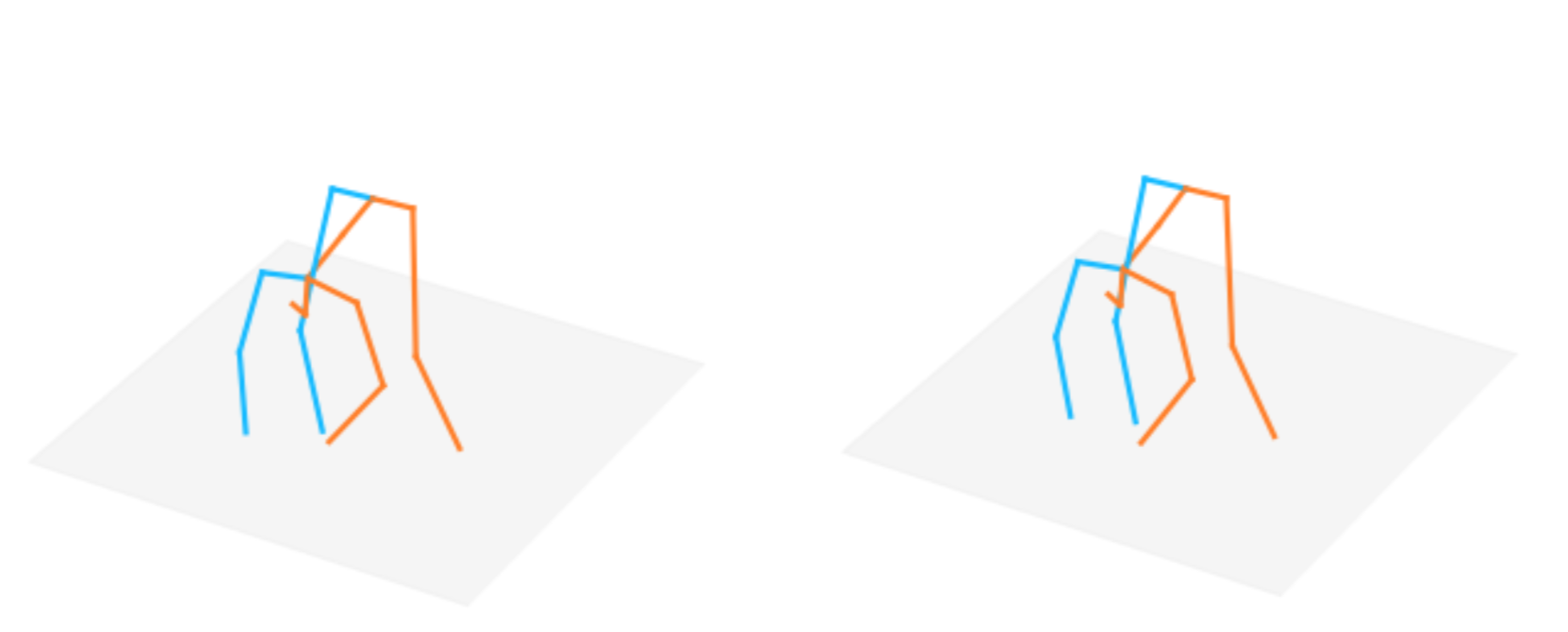}\hspace{\fighspace} & \includegraphics[width=\figsize\textwidth]{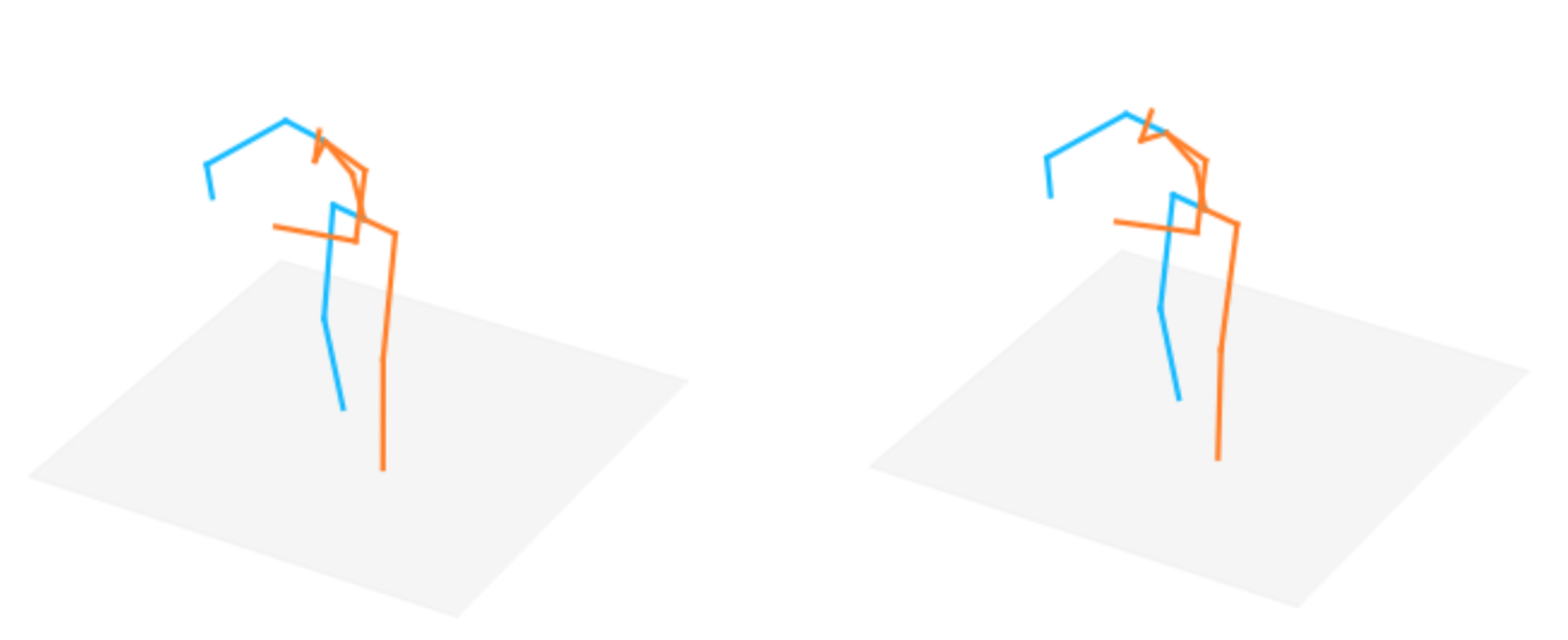}\hspace{\fighspace} & \includegraphics[width=\figsize\textwidth]{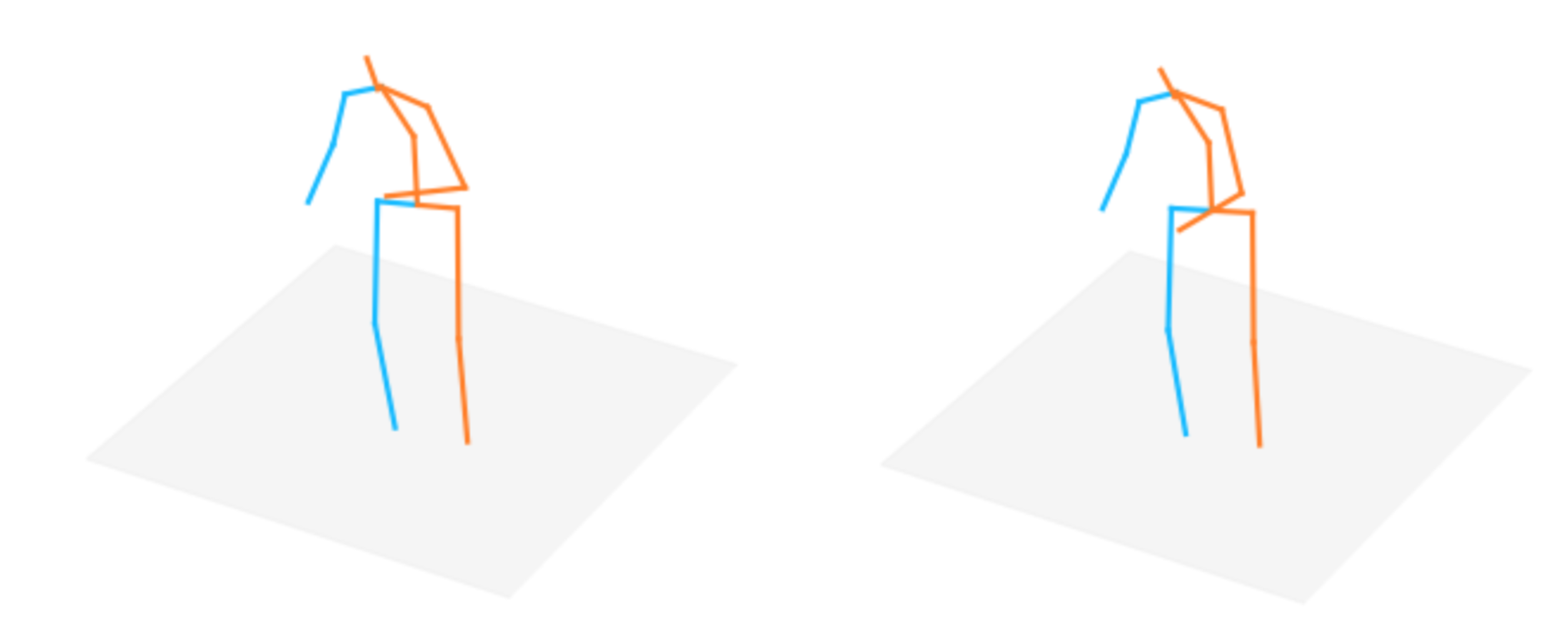}\hspace{\fighspace}  \\

\scriptsize{NP-MPJPE: $0.065$} & \scriptsize{NP-MPJPE: $0.074$}\hspace{\fighspace} & \scriptsize{NP-MPJPE: $0.085$} & \scriptsize{NP-MPJPE: $0.091$} \\

\includegraphics[width=\figsize\textwidth]{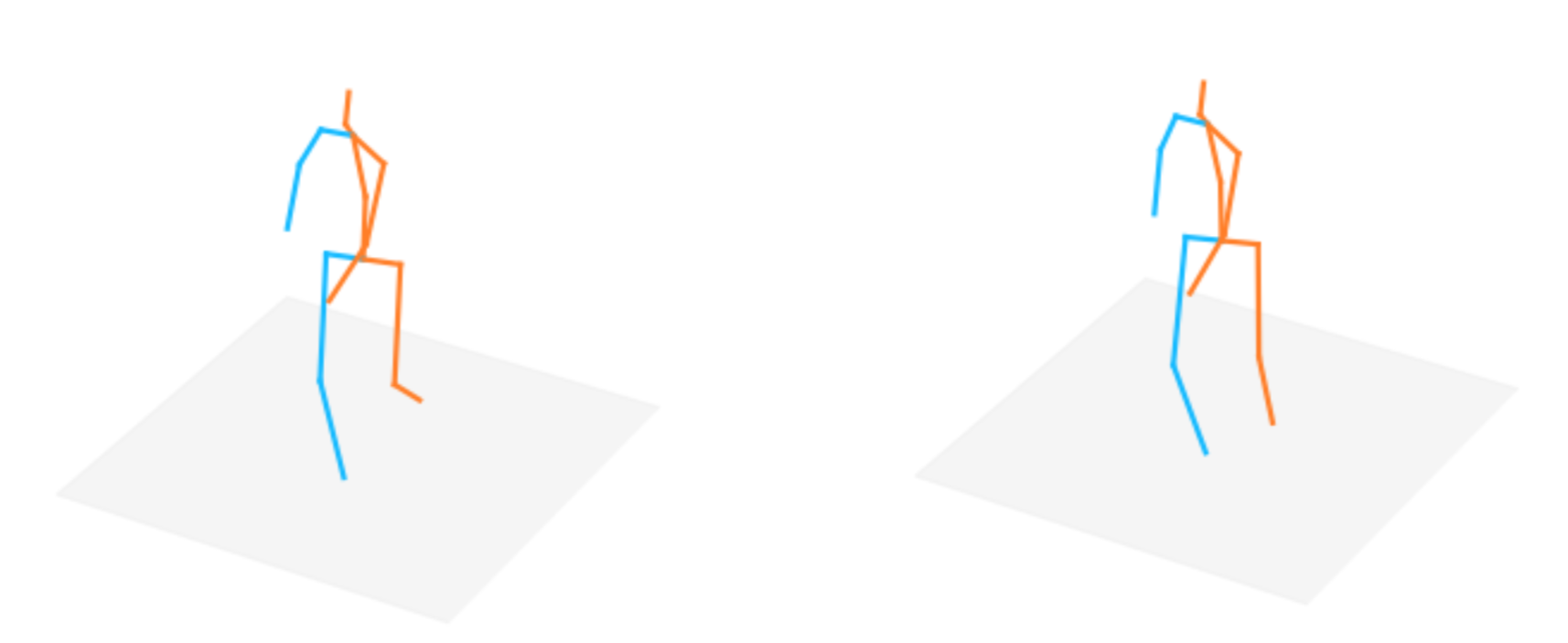}\hspace{\fighspace} & \includegraphics[width=\figsize\textwidth]{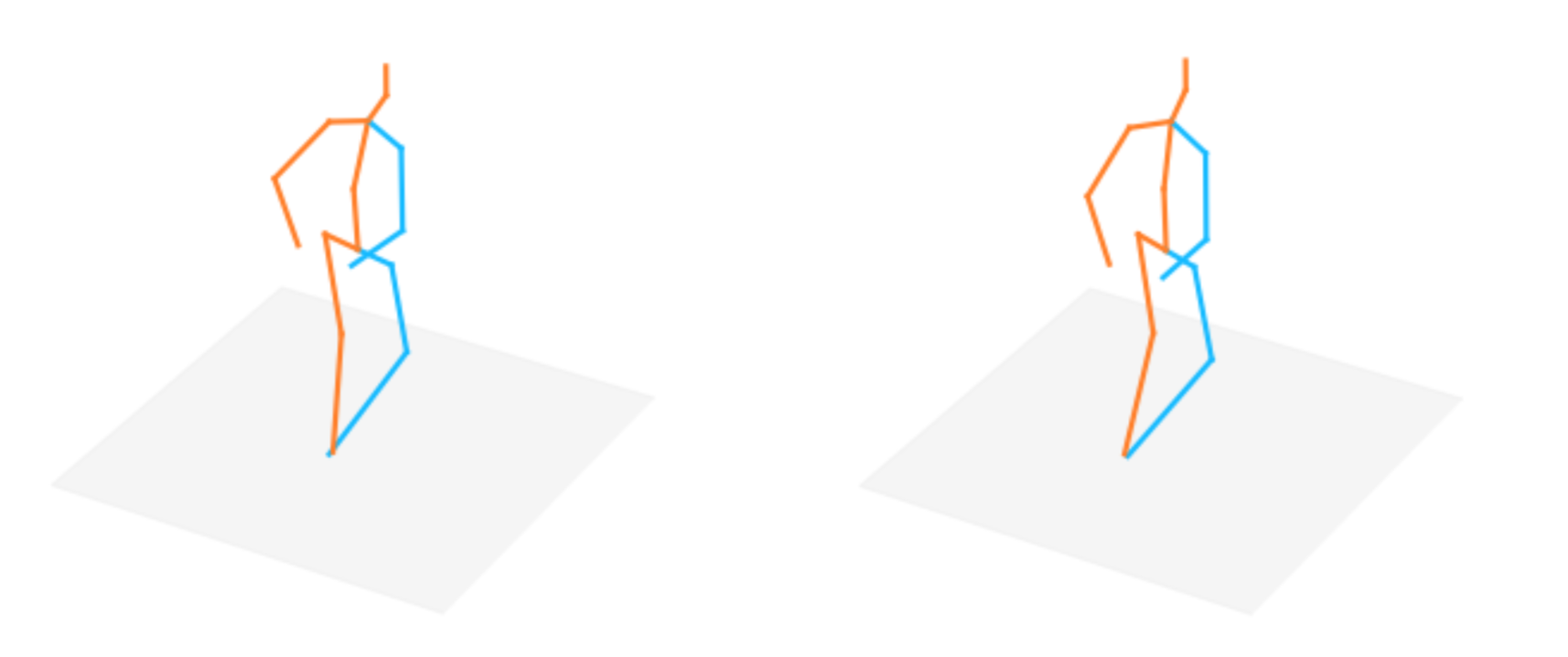}\hspace{\fighspace} & \includegraphics[width=\figsize\textwidth]{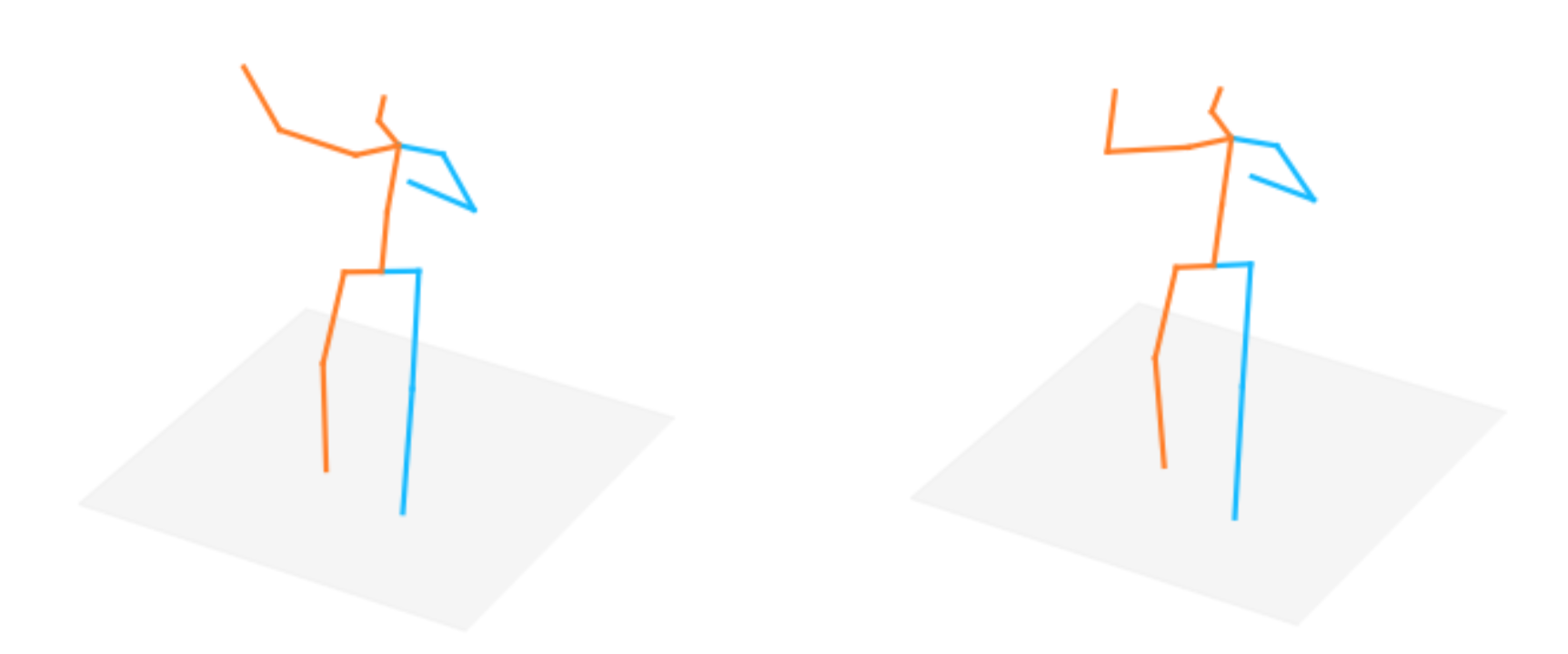}\hspace{\fighspacer}  & \includegraphics[width=\figsize\textwidth]{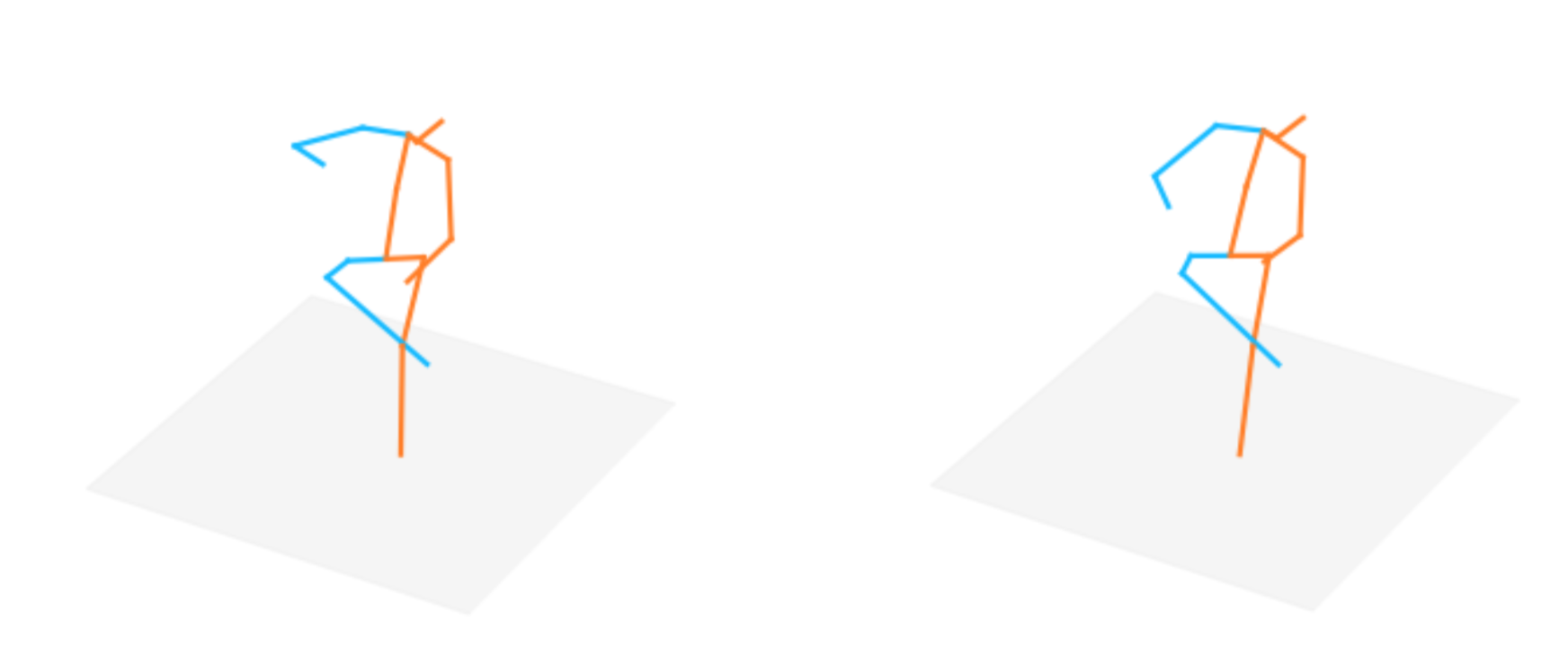}\hspace{\fighspace} \\

\scriptsize{NP-MPJPE: $0.111$} & \scriptsize{NP-MPJPE: $0.121$}\hspace{\fighspace} & \scriptsize{NP-MPJPE: $0.134$} & \scriptsize{NP-MPJPE: $0.145$}\hspace{\fighspace} \\

\includegraphics[width=\figsize\textwidth]{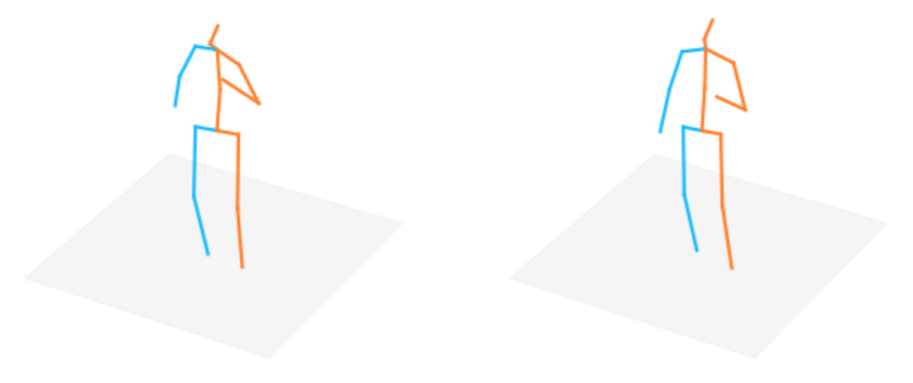}\hspace{\fighspace} & \includegraphics[width=\figsize\textwidth]{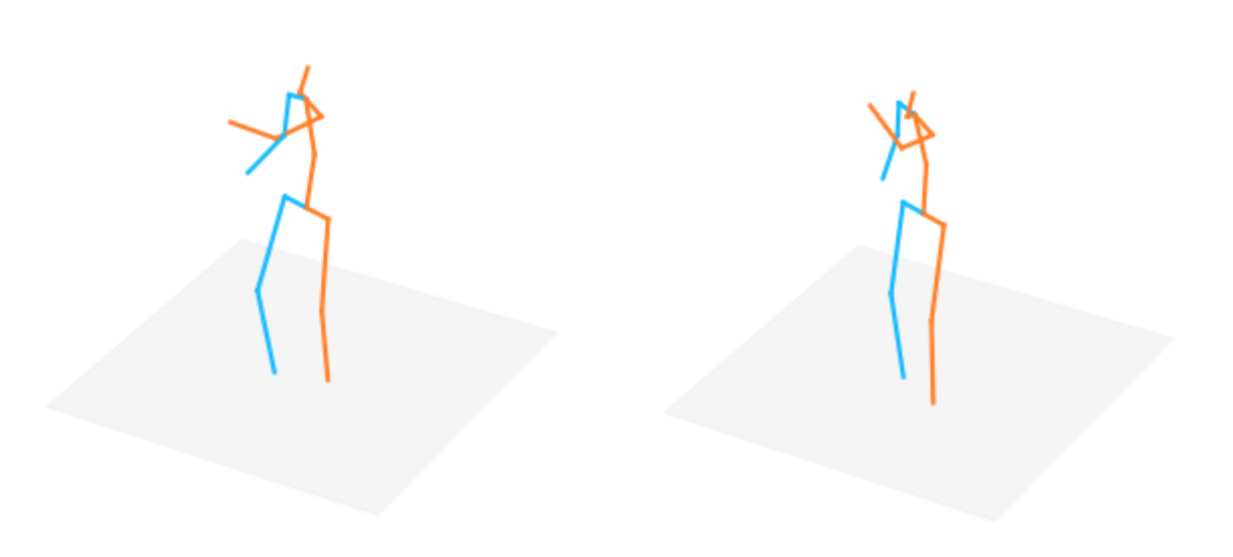}\hspace{\fighspacer} & \includegraphics[width=\figsize\textwidth]{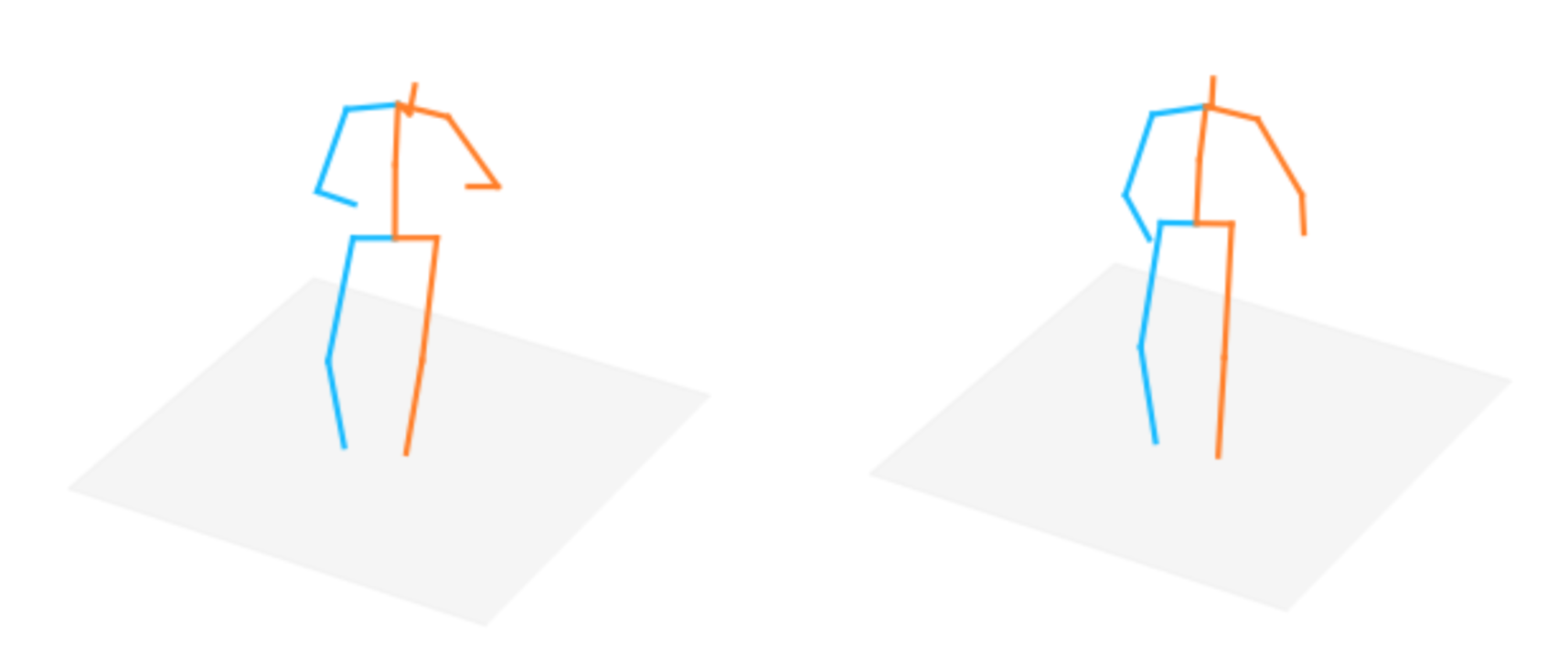}\hspace{\fighspace} & \includegraphics[width=\figsize\textwidth]{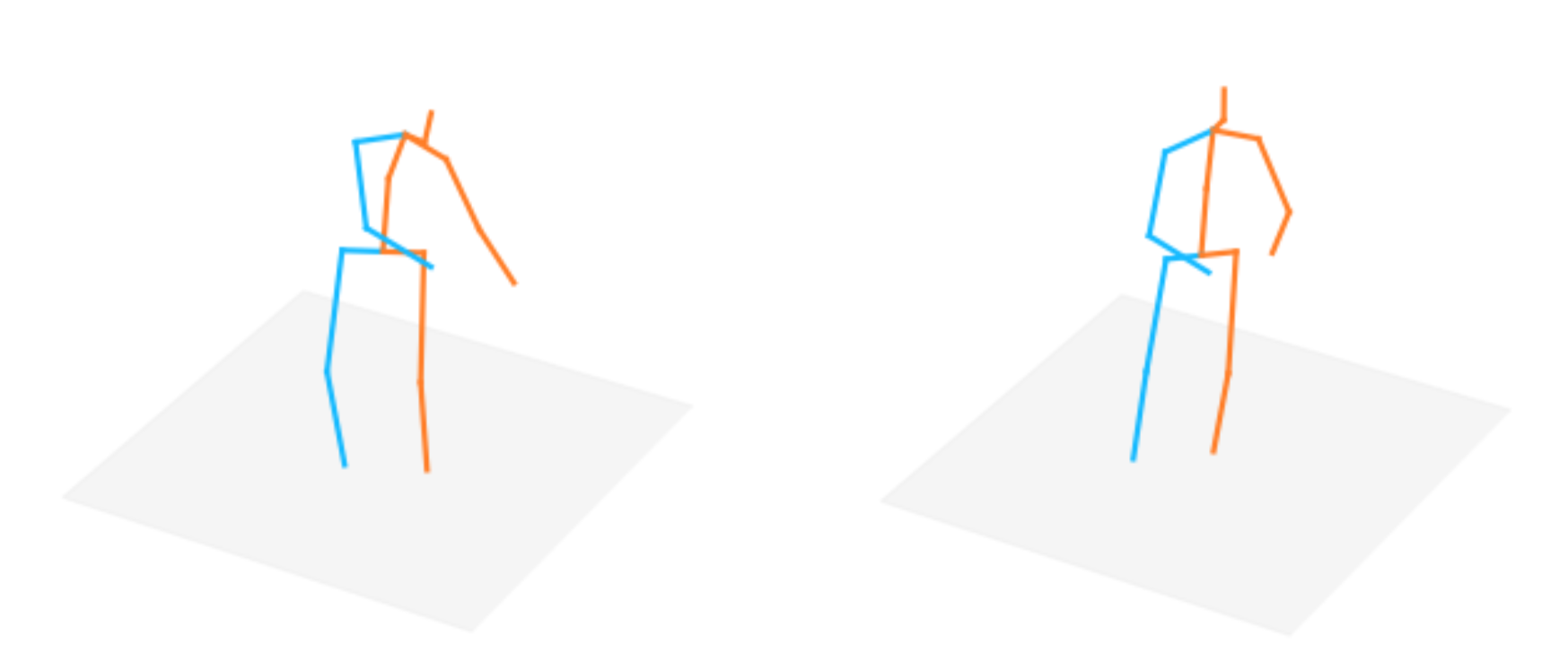}\hspace{\fighspacer} \\

\scriptsize{NP-MPJPE: $0.159$} & \scriptsize{NP-MPJPE: $0.174$}\hspace{\fighspace} & \scriptsize{NP-MPJPE: $0.185$} & \scriptsize{NP-MPJPE: $0.192$}\hspace{\fighspace} \\

\includegraphics[width=\figsize\textwidth]{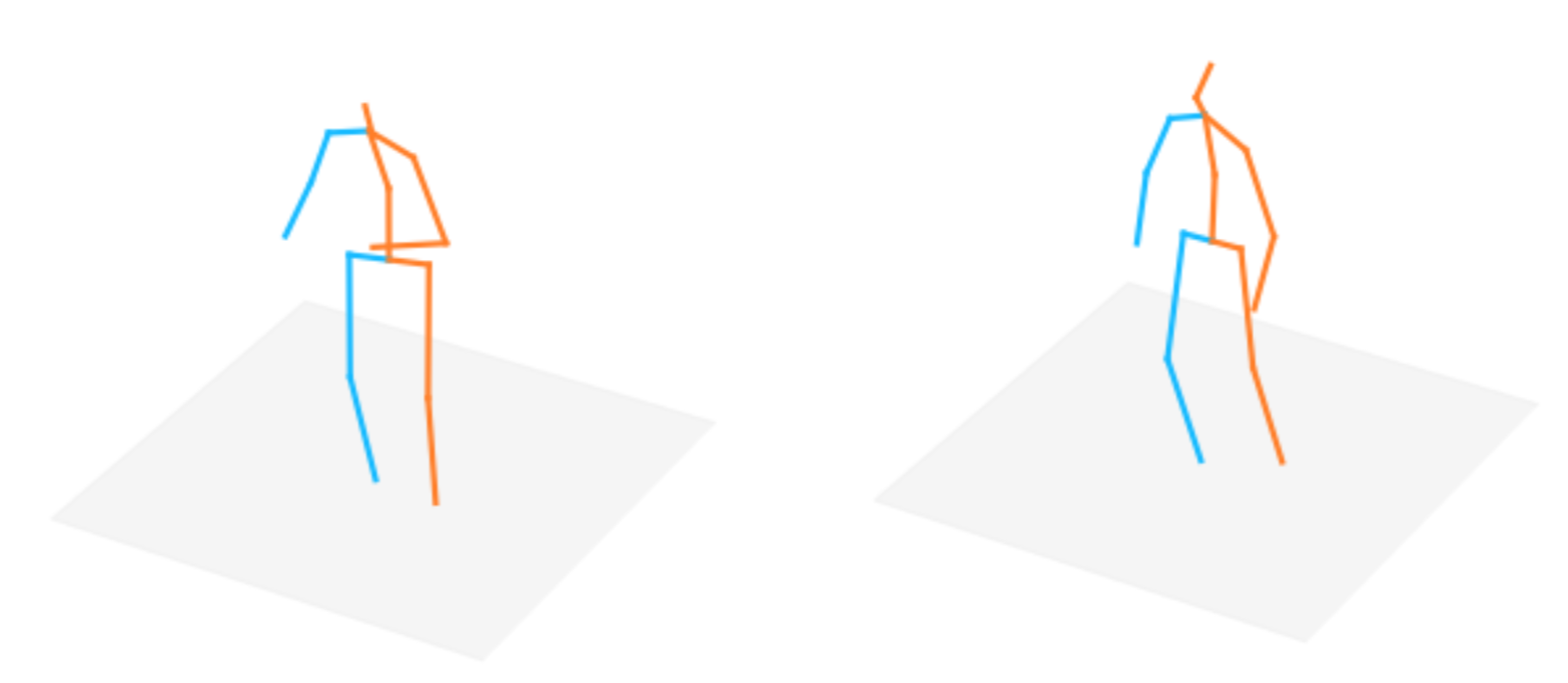}\hspace{\fighspace} & \includegraphics[width=\figsize\textwidth]{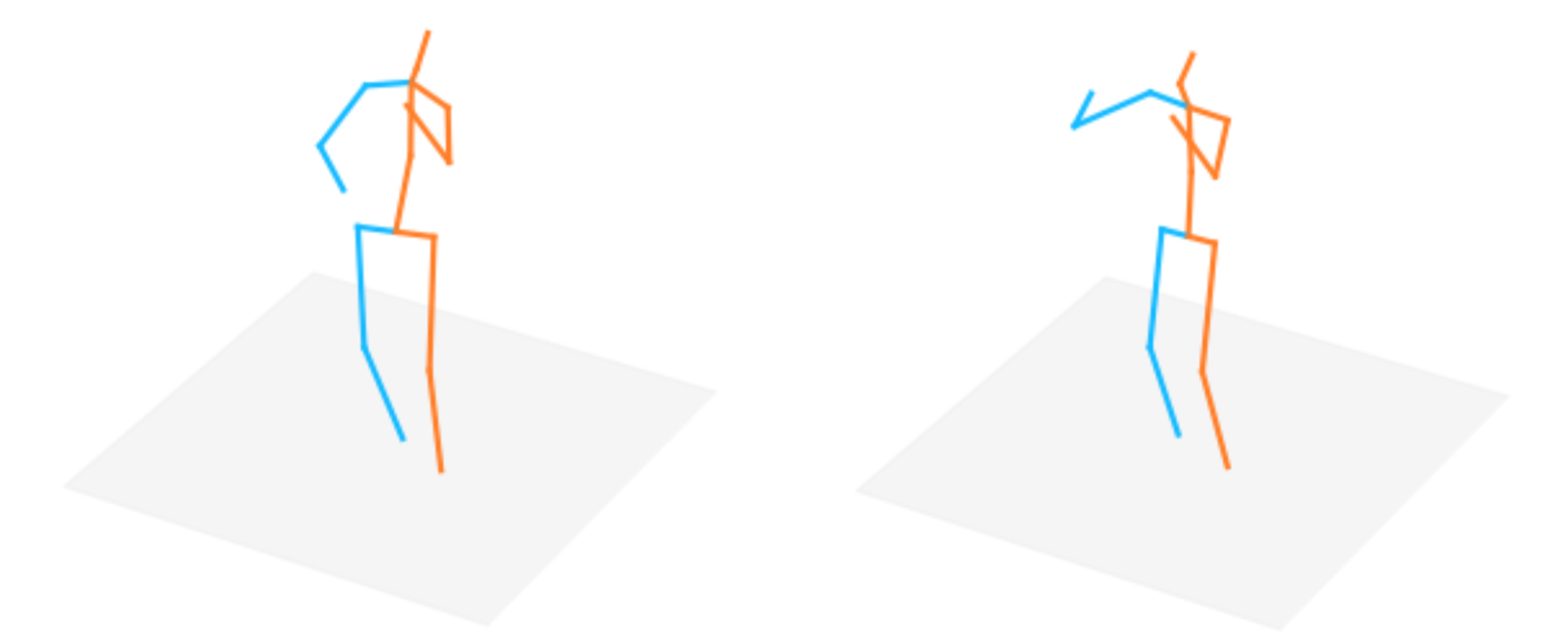}\hspace{\fighspacer} & \includegraphics[width=\figsize\textwidth]{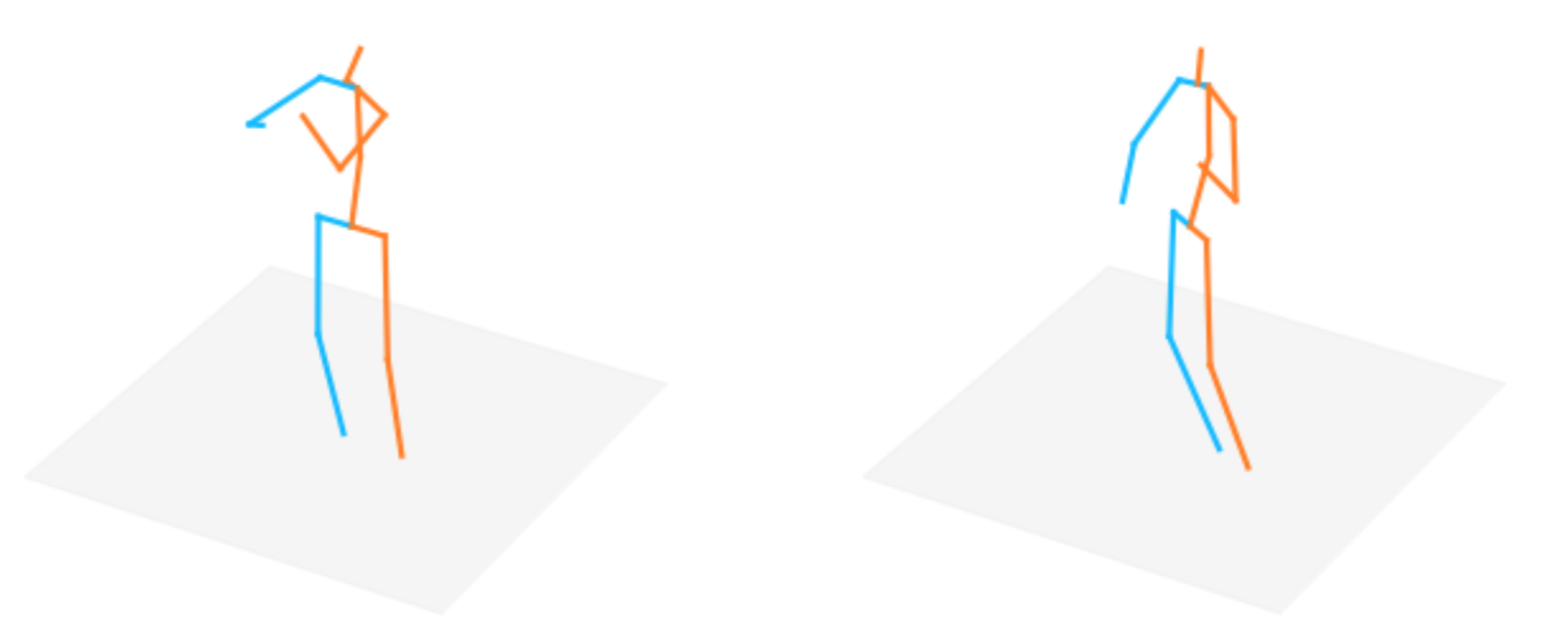}\hspace{\fighspace} & \includegraphics[width=\figsize\textwidth]{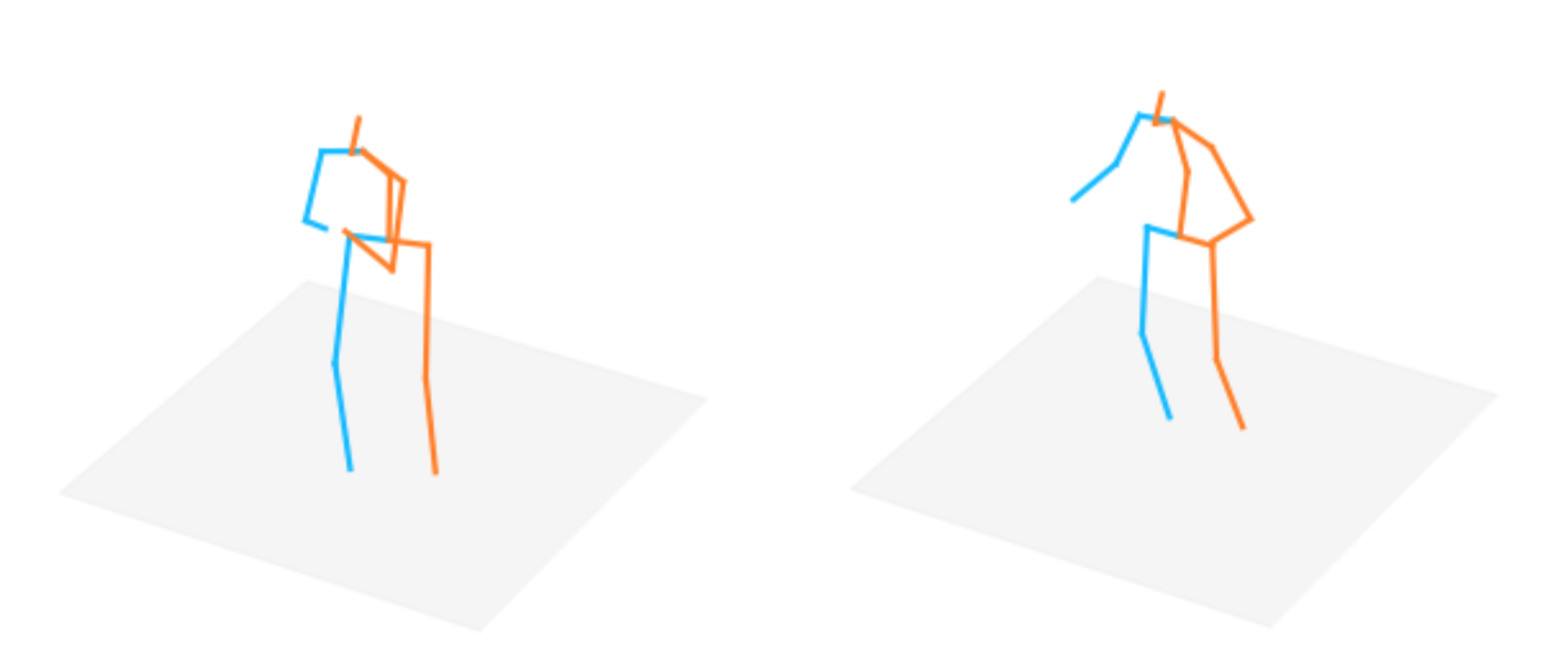}\hspace{\fighspacer} \\

\end{tabular}
\caption{3D pose pairs with different NP-MPJPE, where the NP-MPJPE increases with each row. The poses are randomly sampled from the hold-out set of H3.6M. Row 1 shows pairs with $0.00$ to $0.05$ NP-MPJPE, row 2 shows pairs with $0.05$ to $0.10$ NP-MPJPE, row 3 shows pairs with $0.10$ to $0.15$ NP-MPJPE, and row 4 shows pairs with $0.15$ to $0.20$ NP-MPJPE.}
\label{fig:supp_similarity}
\vspace{-0.2cm}
\end{figure*}

\section{Additional Implementation Details}\label{sec:implementation_details}

The backbone network architecture for our model is based on~\cite{martinez2017simple}. We use two residual blocks, batch normalization, $0.3$ dropout, and no maximum weight norm constraint~\cite{martinez2017simple}. During training, we use exponential moving average with $0.9999$ decay rate and normalize matching probabilities to within $[0.05,0.95]$ for numerical stability. We use Adagrad optimizer~\cite{duchi2011adaptive} with fixed learning rate $0.02$ and batch size $N=256$.

\begin{figure}[t]
  \centering
  \subfloat[17 keypoints based on H3.6M.\label{fig:skeleton17}]{\includegraphics[width=0.45\textwidth]{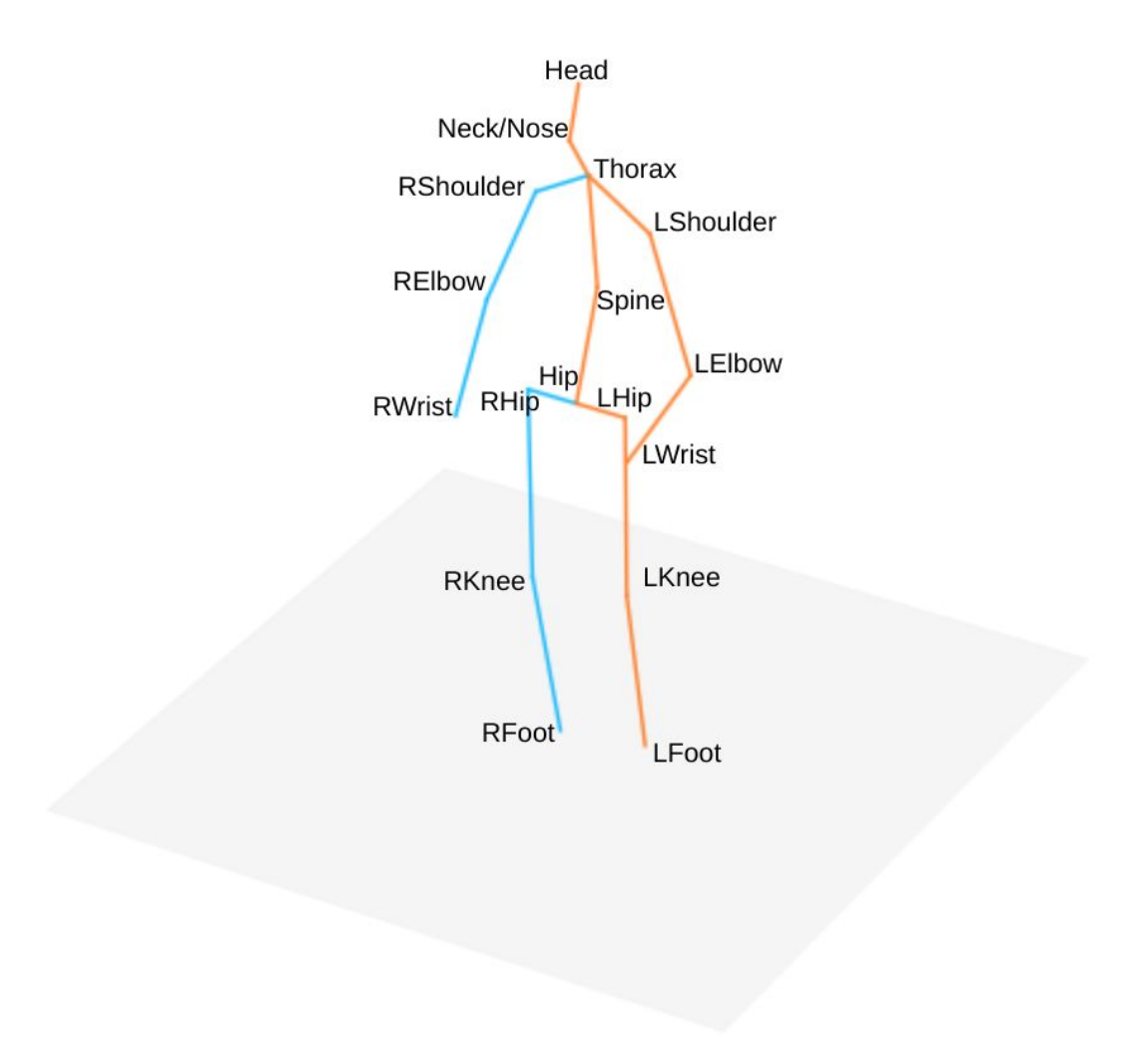}}
  \subfloat[13 keypoints based on COCO.\label{fig:skeleton13}]{\includegraphics[width=0.45\textwidth]{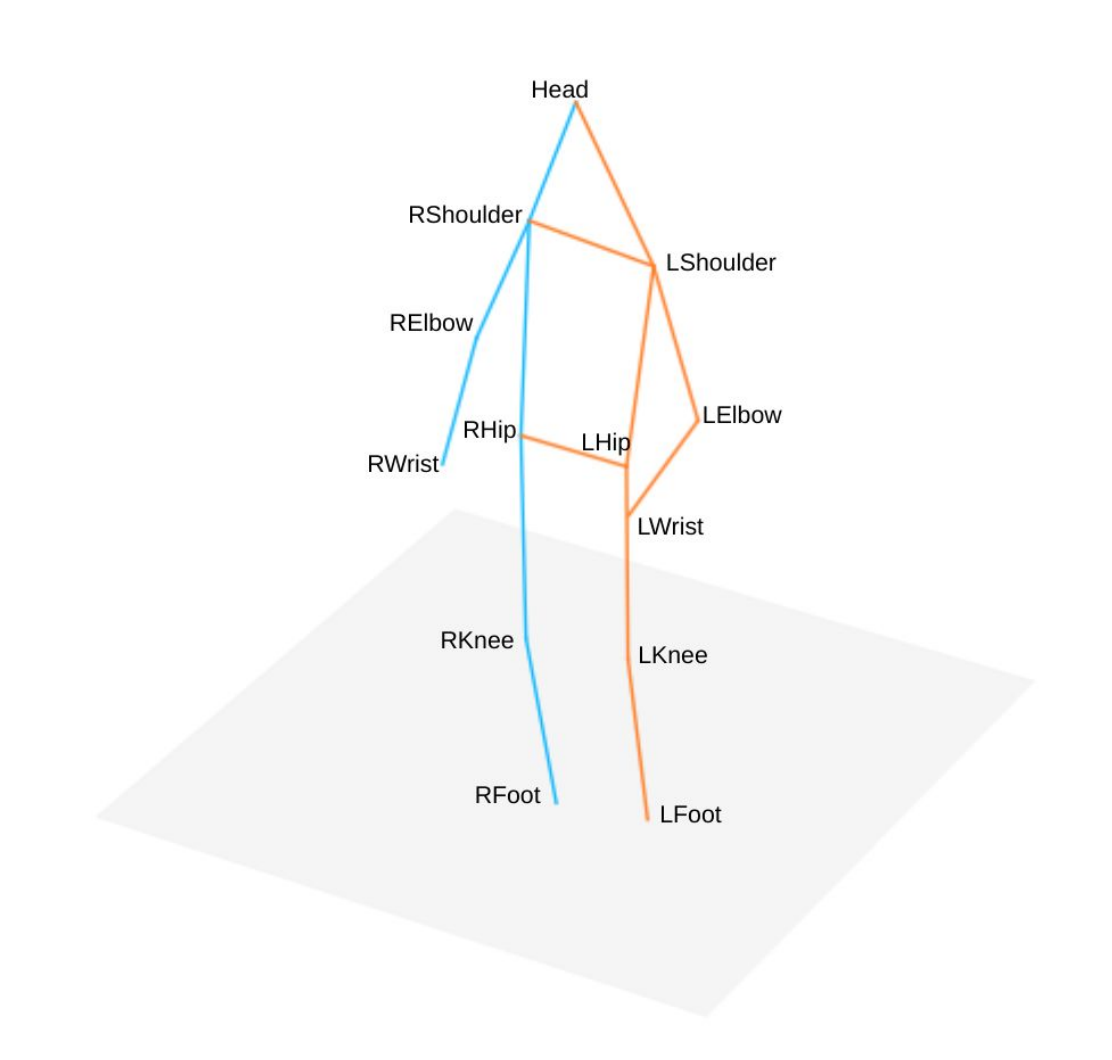}}
  

\caption{Visualization of pose keypoints used in our experiments.}
\label{fig:skeleton}
\end{figure}

\customizedparagraph{Keypoint Definition} Fig.~\ref{fig:skeleton} illustrates the keypoints that we use in our experiments. The 3D poses used in our experiments are the 17 keypoints corresponding to the H3.6M~\cite{ionescu2013human3} skeleton used in~\cite{martinez2017simple}, shown in Fig.~\ref{fig:skeleton17}. We use this keypoint definition to compute NP-MPJPE for 3D poses and evaluate retrieval accuracy. The Pr-VIPE training and inference process do not depend on a particular 2D keypoint detector. Here, we use PersonLab (ResNet152 single-scale)~\cite{papandreou2018personlab} in our experiments. Our 2D keypoints are selected from the keypoints in COCO~\cite{lin2014microsoft}, which is the set of keypoints detected by PersonLab~\cite{papandreou2018personlab}. We use the 12 body keypoints from COCO and select the ``Nose" keypoint as the head, shown in Fig.~\ref{fig:skeleton13}.

\customizedparagraph{Pose Normalization} We normalize our 2D and 3D poses such that camera parameters are not needed during training and inference. 
For 3D poses, our normalization procedure is similar to that in~\cite{chen2019unsupervised}. We translate a 3D pose so that the hip located at the origin. We then scale the hip to spine to thorax distance to a unit scale. 
For 2D poses, we translate the keypoints so that the center between LHip and RHip is at the origin. Then we normalize the pose such that the maximum distance between shoulder and hip joints is $0.5$. This maximum distance is computed between all pairwise distances among RShoulder, LShoulder, RHip, and LHip.

\customizedparagraph{Downstream Task Experiments} For the action recognition experiment, we follow the standard evaluation protocol~\cite{xia2012view} and remove action ``strum guitar'' and several videos in which less than one third of the target person is visible. We use the official train/test split and report the averaged per-class accuracy. For the view-invariant action recognition experiments in which the index set only contains videos from a single view, we exclude the actions that have zero or only one sample under a particular view. We take the bounding boxes provided with the dataset and use~\cite{papandreou2017towards} (ResNet101) for 2D pose keypoint estimation. For frames of which the bounding box is missing, we copy the bounding box from the nearest frame. Finally, since our embedding is chiral, but certain actions can be done with either body side (pitching a baseball with left or right hand), when we compare two frames, we extract our embeddings from both the original and the mirrored version of each frame, and use the minimum distance between all the pairwise combinations as the frame distance.

For the video alignment experiment, we follow the protocol in~\cite{dwibedi2019temporal}, excluding ``jump rope'' and ``strum guitar'' from our evaluation. For the evaluations between videos under only the same or different views, we exclude actions that have zero videos under a particular view from the average Kendall's Tau computation. Since certain actions can be done with either body side, for a video pair $(v_1,v_2)$, we compute the Kendall's Taus between $(v_1,v_2)$ and $(v_1,\text{mirror}(v_2))$, and use the larger number.

\section{Additional Ablation Studies}\label{sec:ablation}

\customizedparagraph{Effect of Number of Samples $K$ and Margin Parameter $\beta$} Table~\ref{tab:ablation} shows the effect of the number of samples $K$ and the margin parameter $\beta$ (actual triplet margin $\alpha=\log\beta$) on Pr-VIPE. The number of samples control how many points we sample from the embedding distribution to compute matching probability and $\beta$ controls the ratio of matching probability between matching and non-matching pairs. Our model is robust to the choice of $\beta$ in terms of retrieval accuracy as shown by Table~\ref{tab:ablation}. The main effect of $\beta$ is on retrieval confidence, as non-matching pairs are scaled to a smaller matching probability for larger $\beta$. Pr-VIPE performance with 10 samples is competitive with the baselines in the main paper, but we do better with 20 samples. Increasing the number of samples further has similar performance. For our experiments, we use $20$ samples and $\beta = 2$.

\begin{table}
  \centering
  \caption{Additional ablation study results of Pr-VIPE on H3.6M with the number of samples $K$ and margin parameter $\beta$.} \label{tab:ablation}
  \scalebox{1.00}{
   \begin{tabular}{c c | c c c}
   \toprule[0.2em]
    Hyperparameter & Value & Hit@$1$ & Hit@$10$ & Hit@$20$ \\ [0.25ex]
   \toprule[0.2em]
 \multirow{3}{*}{$K$} & $10$ & $0.744$ & $0.948$ & $0.971$\\
  & $20$ & $0.762$ & $0.956$ & $0.977$ \\
  & $30$ & $0.755$ & $0.955$ & $0.975$\\
 \hline
 \multirow{4}{*}{$\beta$} & $1.25$ & $0.758$ & $0.956$ & $0.977$\\
  & $1.5$ & $0.759$ & $0.956$ & $0.977$ \\
  & $2$ & $0.762$ & $0.956$ & $0.977$ \\
  & $3$ & $0.760$ & $0.955$ & $0.976$\\
   \bottomrule[0.1em]
\end{tabular}
}
\end{table}

\customizedparagraph{Effect of Camera Augmentation} We explore the effect of different random rotations during camera augmentation on pose retrieval results in Table~\ref{tab:ablation_aug}. All models are trained on the 4 chest-level cameras on H3.6M but the models with camera augmentation also use projected 2D keypoints from randomly rotated 3D poses. For the random rotation, we always use azimuth range of $\pm180^{\circ}$, and we test performance with different angle limits for elevation and roll. We see that the model with no augmentation does the best on the H3.6M, which has the same 4 camera views as training. With increase in rotation angles during mixing, the performance on chest-level cameras drop while performance on new camera views generally increases. The results demonstrate that mixing detected and projected keypoints reduces model overfitting on camera views used during training. Training using randomly rotated keypoints enables our model to generalize much better to new views.

\begin{table}
  \centering
  \caption{Additional ablation study results of Pr-VIPE on H3.6M and 3DHP using different rotation thresholds for camera augmentation. The angle threshold for azimuth is always $\pm180^{\circ}$ and the angle thresholds in the table are for elevation and roll. The row for w/o aug. corresponds to Pr-VIPE without augmentation.} \label{tab:ablation_aug}
  \scalebox{1.00}{
   \begin{tabular}{c c | c c c c c}
   \toprule[0.2em]
   & & \multicolumn{3}{c}{Hit@$1$ on evaluation dataset}\\
    Hyperparameter & Range & H3.6M & 3DHP (all) & 3DHP (chest) \\ [0.25ex]
   \toprule[0.2em]
  \multirow{4}{*}{Elevation and Roll Angle} & w/o aug. & $0.762$ & $0.199$ & $0.255$\\
  & $\pm15^{\circ}$ & $0.747$ & $0.252$ & $0.289$ \\
  & $\pm30^{\circ}$ & $0.737$ & $0.264$ & $0.283$\\
  & $\pm45^{\circ}$ & $0.737$ & $0.262$ & $0.273$ \\
   \bottomrule[0.1em]
\end{tabular}
}
\end{table}

\begin{table}
  \centering
  \caption{Additional ablation study results of Pr-VIPE on H3.6M with different NP-MPJPE threshold $\kappa$ for training and evaluation.} \label{tab:ablation_kappa}
  \scalebox{1.00}{
  \begin{tabular}{c | c c c c}
  \toprule[0.2em]
  & \multicolumn{4}{c}{Hit@$1$ with evaluation $\kappa$}\\
  Training $\kappa$ & $0.05$ & $0.10$ & $0.15$ & $0.20$ \\[0.25ex]
  \toprule[0.2em]
 $0.05$ & $\bm{0.495}$ & $0.761$ & $0.908$ & $0.962$\\
 $0.10$ & $0.489$ & $\bm{0.762}$ & $0.909$ & $0.963$ \\
 $0.15$ & $0.462$ & $0.753$ & $\bm{0.910}$ & $\bm{0.965}$\\
 $0.20$ & $0.429$ & $0.731$ & $0.906$ & $\bm{0.965}$\\
  \bottomrule[0.1em]
\end{tabular}
}
\end{table}

\customizedparagraph{Effect of NP-MPJPE threshold $\kappa$} We train and evaluate with different values of the NP-MPJPE threshold $\kappa$ in Table~\ref{tab:ablation_kappa}. $\kappa$ controls the NP-MPJPE threshold for a matching pose pair and visualizations of pose pairs with different NP-MPJPE are in Fig.~\ref{fig:supp_similarity}. Table~\ref{tab:ablation_kappa} shows that Pr-VIPE generally achieves the best accuracy for a given NP-MPJPE threshold when the model is trained with the same matching threshold. Additionally, when we train with a tight threshold, e.g., $\kappa = 0.05$, we do comparatively well on accuracy at looser thresholds. In contrast, when we train with a loose threshold, e.g., $\kappa = 0.20$, we do not do as well given a tighter accuracy threshold at evaluation. This is because when we push non-matching poses using the triplet ratio loss, $\kappa = 0.20$ only pushes poses that are more than $0.20$ NP-MPJPE apart, and does not explicitly push poses less than the NP-MPJPE threshold. The closest retrieved pose will then be within $0.20$ NP-MPJPE but it is not guaranteed to be within any threshold $<0.20$ NP-MPJPE. But when we use $\kappa = 0.05$ for training, poses that are more than $0.05$ NP-MPJPE are pushed apart, which also satisfies $\kappa = 0.20$ threshold.

In the main paper, we use $\kappa = 0.1$. For future applications with other matching definitions, the Pr-VIPE framework is flexible and can be trained with different $\kappa$ to satisfy different accuracy requirements.

\begin{figure*}[t!]
  \centering
  \subfloat[Poses with top-$3$ smallest variance and their nearest neighbors in terms of 2D NP-MPJPE.\label{fig:smallest_variance}]{\includegraphics[width=\textwidth]{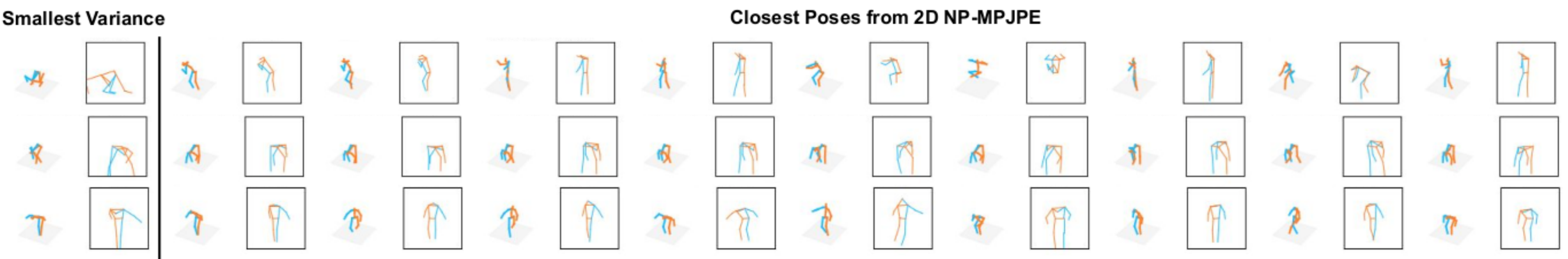}}\\
  \subfloat[Poses with top-$3$ largest variance and their nearest neighbors in terms of 2D NP-MPJPE.\label{fig:largest_variance}]{\includegraphics[width=\textwidth]{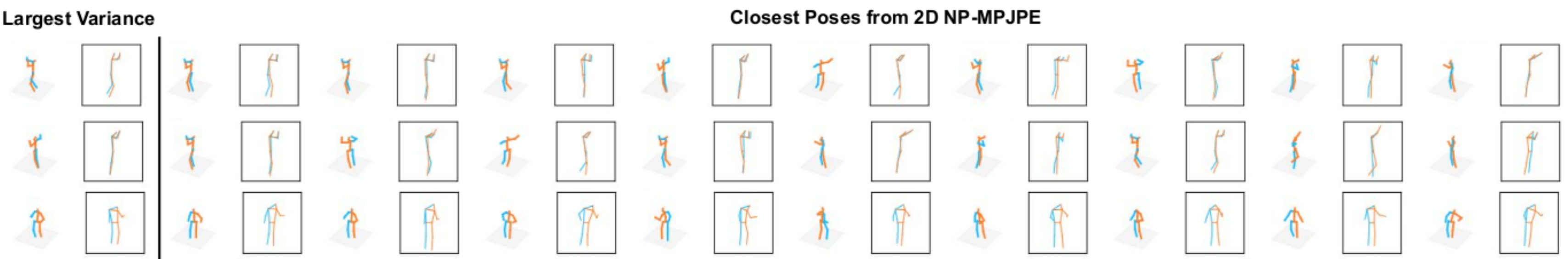}}
  


\caption{Top retrievals by 2D NP-MPJPE from H3.6M hold-out subset for queries with top-$3$ largest and smallest variances. 2D poses are shown in the boxes.}
\label{fig:variance}
\end{figure*}

\customizedparagraph{Additional Plots for Ordered Variances} Similar to the main paper, we retrieve poses using 2D NP-MPJPE for the top-$3$ 2D poses with smallest and largest variances in Fig.~\ref{fig:variance}. Fig.~\ref{fig:smallest_variance} shows that for the poses with the top-$3$ smallest variances, the nearest 2D pose neighbors are visually distinct, which means that these 2D poses are less ambiguous. On the other hand, the nearest 2D pose neighbors of the poses with the largest variances in Fig.~\ref{fig:largest_variance} are visually similar, which means that these 2D poses are more ambiguous.

\begin{figure*}[t!]
  \centering
  \includegraphics[width=0.6\linewidth]{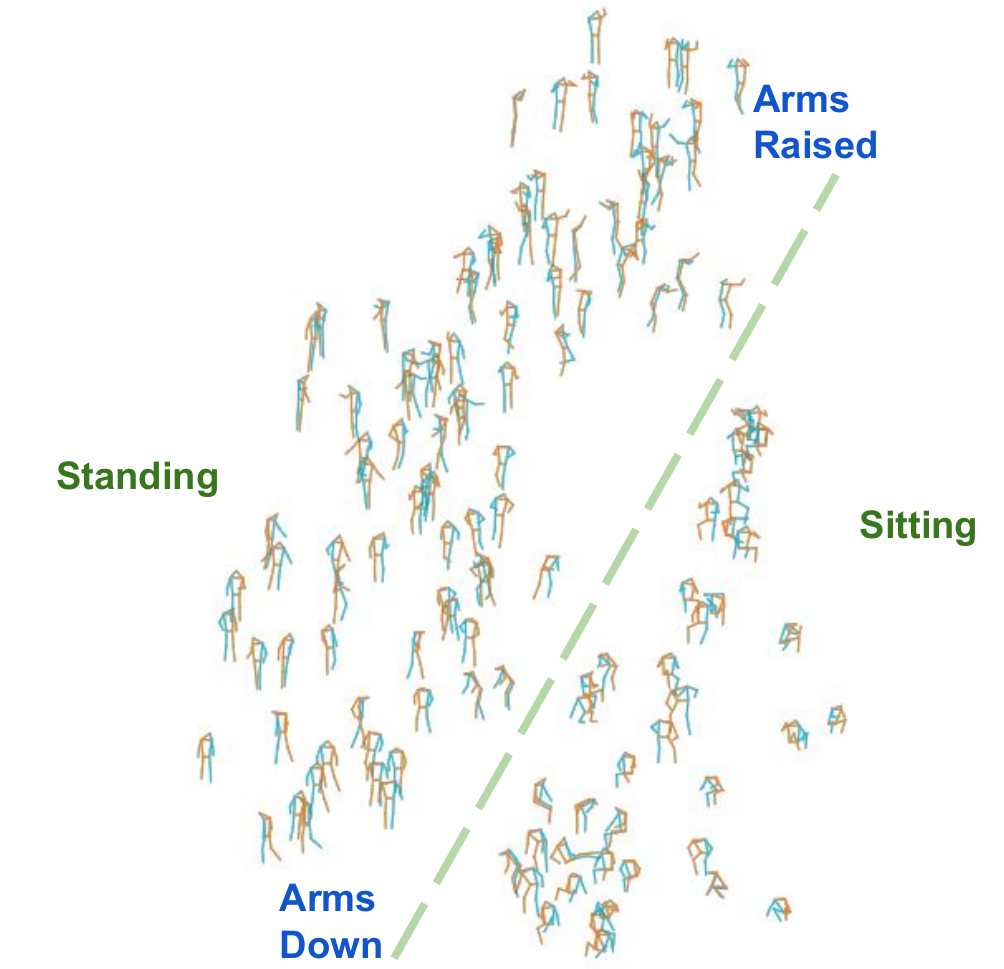}
  \caption{Visualization of Pr-VIPE space with 2D poses in the H3.6M hold-out subset using the first two PCA dimensions.}
  \label{fig:pca_supp}
  \vspace{-0.3cm}
\end{figure*}

\customizedparagraph{Embedding Space Visualization} We run Principal Component Analysis (PCA) on the $16$-dimensional embeddings using the Pr-VIPE model. Fig.~\ref{fig:pca_supp} visualizes the first two principal dimensions. To visualize more unique poses, we randomly subsample the H3.6M hold-out set and select 3D poses at least 0.1 NP-MPJPE apart. Fig.~\ref{fig:pca_supp} demonstrates that 2D poses from similar 3D poses are close together, while non-matching poses are further apart. Standing and sitting poses seem well separated from the two principle dimensions. Additionally, there are leaning poses between sitting and standing. Poses near the top of the figure have arms raised, and there is generally a gradual transition to the bottom of the figure, where arms are lowered. These results show that from 2D joint keypoints only, we are able to learn view-invariant properties with compact embeddings.

\section{Additional Quantitative Pose Retrieval Results}\label{sec:additional_comp}

We show an additional view-invariant pose retrieval evaluation comparing Pr-VIPE (with camera augmentation) to EpipolarPose~\cite{kocabas2019self}, a recent multi-view image based model, on cross-view pose retrieval. For Human3.6M, EpipolarPose is trained with the same training split as Pr-VIPE. The evaluation split we use is a frame subset provided by~\cite{kocabas2019self} for which the authors provided cropping boxes based on groundtruth 3D keypoints. The input images are cropped using these bounding boxes, and the trained models provided by the authors are then ran on the cropped images. In this way, we evaluate EpipolarPose using all the information provided by the authors. In comparison, Pr-VIPE uses detected keypoints and no groundtruth information for inference.

We show retrieval results on Human3.6M since~\cite{kocabas2019self} is based on images and requires a different model to be trained for 3DHP. We emphasize that this is a different evaluation split from our main paper, since we use the evaluation subset of Human3.6M for which~\cite{kocabas2019self} provides bounding boxes. On this subset, Pr-VIPE with augmentation achieves $75.2\%$ Hit@$1$, fully supervised EpipolarPose achieves $72.7\%$ Hit@$1$ and self-supervised EpipolarPose achieves $67.8\%$ Hit@$1$.

These results show the effectiveness of Pr-VIPE for pose retrieval. Our model, using detected 2D keypoints and no groundtruth information, can retrieve poses more accurately compared with~\cite{kocabas2019self}. We further note that 3D pose estimation models require rigid alignment between every query-index pairs to achieve their best performance for retrieval, while Pr-VIPE does not require post-processing. 

\section{Additional Qualitative Pose Retrieval Results}\label{sec:pose_retrieval_qres}

We present more view-invariant pose retrieval qualitative results for Pr-VIPE on all the relevant datasets in Fig.~\ref{fig:supp_retrieval}. The first two rows show results on H3.6M, the next three rows are on 3DHP and the last two rows shows results using the hold-out set in H3.6M to retrieve from 2DHP. We are able to retrieve across camera views and subjects on all datasets.

\def\figsize{0.15}
\def\fighspace{-0.5mm}
\def\fighspacer{2.5mm}
\begin{figure*}
\centering
\begin{tabular}{cccccc}
\centering
\scriptsize{$C=0.990$}\hspace{\fighspace} & \scriptsize{$E=0.084$}\hspace{\fighspacer} & \scriptsize{$C=0.983$}\hspace{\fighspace} & \scriptsize{$E=0.000$}\hspace{\fighspacer} & \scriptsize{$C=0.994$}\hspace{\fighspace} & \scriptsize{$E=0.132$}\hspace{\fighspacer} \\
\includegraphics[width=\figsize\textwidth]{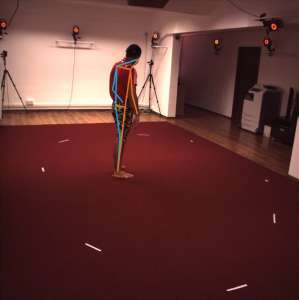}\hspace{\fighspace} & \includegraphics[width=\figsize\textwidth]{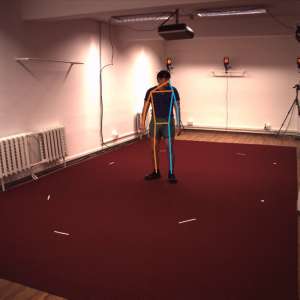}\hspace{\fighspacer}  & \includegraphics[width=\figsize\textwidth]{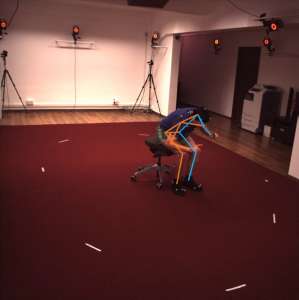}\hspace{\fighspace} & \includegraphics[width=\figsize\textwidth]{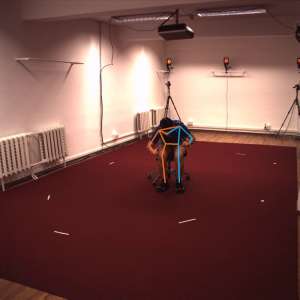}\hspace{\fighspacer} &
\includegraphics[width=\figsize\textwidth]{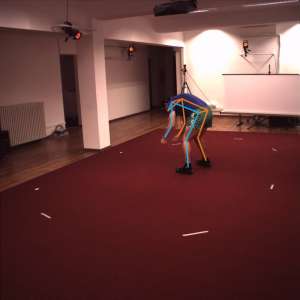}\hspace{\fighspace} & \includegraphics[width=\figsize\textwidth]{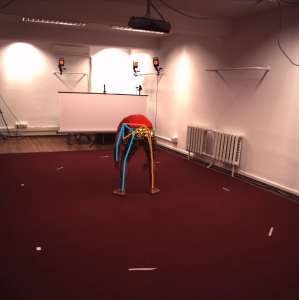}\hspace{\fighspacer} \\

\scriptsize{$C=0.905$}\hspace{\fighspace} & \scriptsize{$E=0.066$}\hspace{\fighspacer} & \scriptsize{$C=0.905$}\hspace{\fighspace} & \scriptsize{$E=0.094$}\hspace{\fighspacer} & \scriptsize{$C=0.888$}\hspace{\fighspace} & \scriptsize{$E=0.130$}\hspace{\fighspacer} \\
\includegraphics[width=\figsize\textwidth]{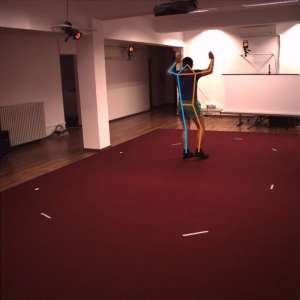}\hspace{\fighspace} & \includegraphics[width=\figsize\textwidth]{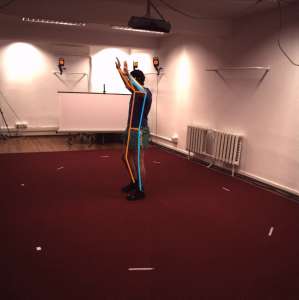}\hspace{\fighspacer}  & \includegraphics[width=\figsize\textwidth]{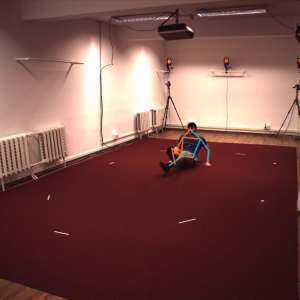}\hspace{\fighspace} & \includegraphics[width=\figsize\textwidth]{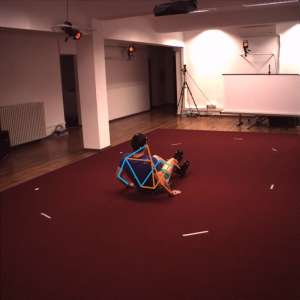}\hspace{\fighspacer} &
\includegraphics[width=\figsize\textwidth]{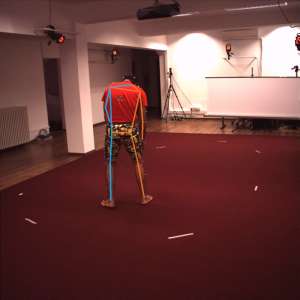}\hspace{\fighspace} & \includegraphics[width=\figsize\textwidth]{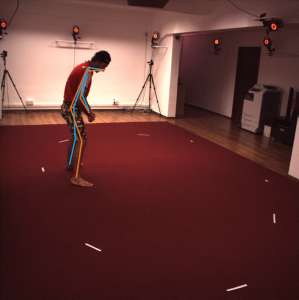}\hspace{\fighspacer} \\

\scriptsize{$C=0.806$}\hspace{\fighspace} & \scriptsize{$E=0.085$}\hspace{\fighspacer} & \scriptsize{$C=0.891$}\hspace{\fighspace} & \scriptsize{$E=0.119$}\hspace{\fighspacer} & \scriptsize{$C=0.599$}\hspace{\fighspace} & \scriptsize{$E=0.222$}\hspace{\fighspacer} \\
\includegraphics[width=\figsize\textwidth]{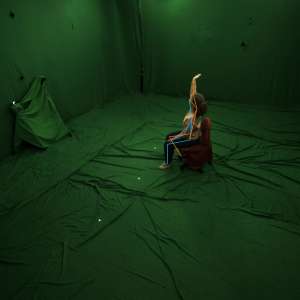}\hspace{\fighspace} & \includegraphics[width=\figsize\textwidth]{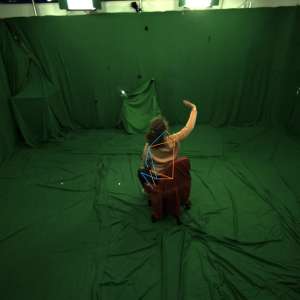}\hspace{\fighspacer}  & \includegraphics[width=\figsize\textwidth]{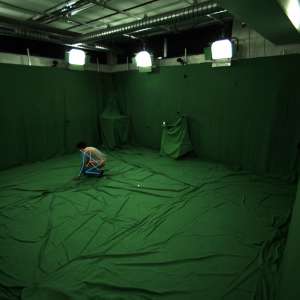}\hspace{\fighspace} & \includegraphics[width=\figsize\textwidth]{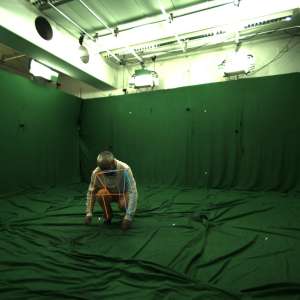}\hspace{\fighspacer} &
\includegraphics[width=\figsize\textwidth]{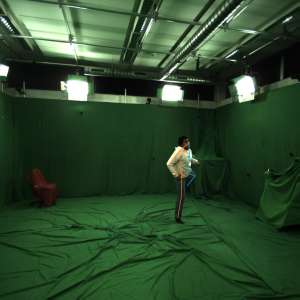}\hspace{\fighspace} & \includegraphics[width=\figsize\textwidth]{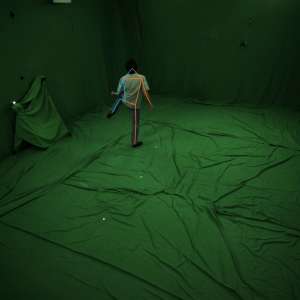}\hspace{\fighspacer} \\

\scriptsize{$C=0.999$}\hspace{\fighspace} & \scriptsize{$E=0.048$}\hspace{\fighspacer} & \scriptsize{$C=0.976$}\hspace{\fighspace} & \scriptsize{$E=0.152$}\hspace{\fighspacer} & \scriptsize{$C=0.727$}\hspace{\fighspace} & \scriptsize{$E=0.295$}\hspace{\fighspacer} \\
\includegraphics[width=\figsize\textwidth]{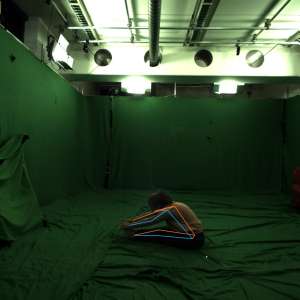}\hspace{\fighspace} & \includegraphics[width=\figsize\textwidth]{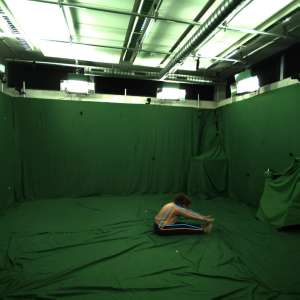}\hspace{\fighspacer}  & \includegraphics[width=\figsize\textwidth]{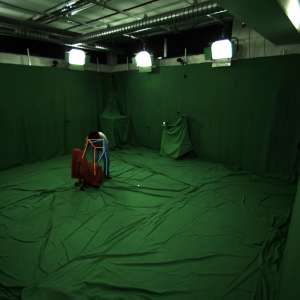}\hspace{\fighspace} & \includegraphics[width=\figsize\textwidth]{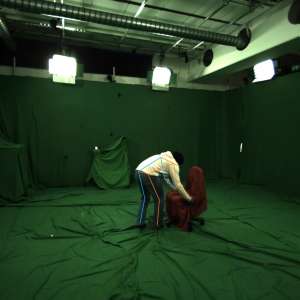}\hspace{\fighspacer} &
\includegraphics[width=\figsize\textwidth]{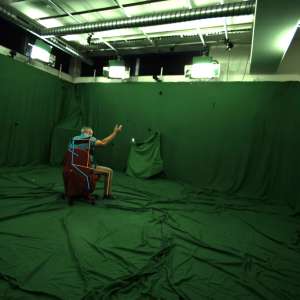}\hspace{\fighspace} & \includegraphics[width=\figsize\textwidth]{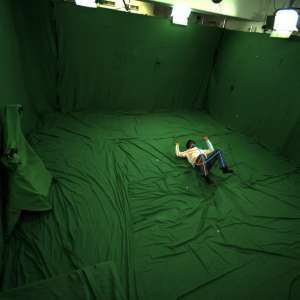}\hspace{\fighspacer} \\

\scriptsize{$C=0.780$}\hspace{\fighspace} & \scriptsize{$E=0.085$}\hspace{\fighspacer} & \scriptsize{$C=0.951$}\hspace{\fighspace} & \scriptsize{$E=0.208$}\hspace{\fighspacer} & \scriptsize{$C=0.918$}\hspace{\fighspace} & \scriptsize{$E=0.328$}\hspace{\fighspacer} \\
\includegraphics[width=\figsize\textwidth]{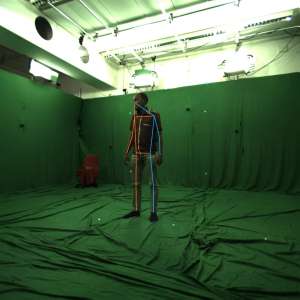}\hspace{\fighspace} & \includegraphics[width=\figsize\textwidth]{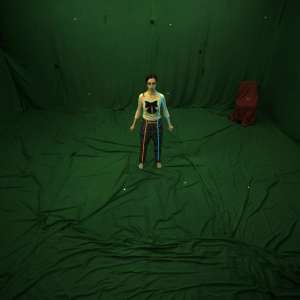}\hspace{\fighspacer}  & \includegraphics[width=\figsize\textwidth]{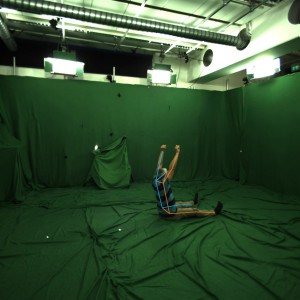}\hspace{\fighspace} & \includegraphics[width=\figsize\textwidth]{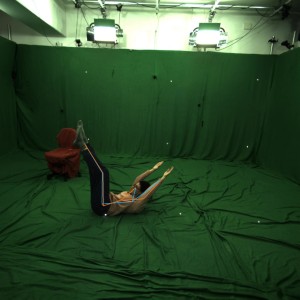}\hspace{\fighspacer} &
\includegraphics[width=\figsize\textwidth]{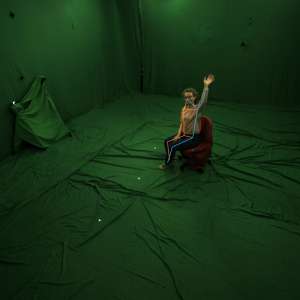}\hspace{\fighspace} & \includegraphics[width=\figsize\textwidth]{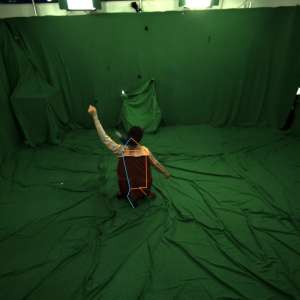}\hspace{\fighspacer} \\

\scriptsize{$C=0.867$}\hspace{\fighspace} & \hspace{\fighspacer} & \scriptsize{$C=0.321$}\hspace{\fighspace} & \hspace{\fighspacer} & \scriptsize{$C=0.936$}\hspace{\fighspace} & \hspace{\fighspacer} \\
\includegraphics[width=\figsize\textwidth]{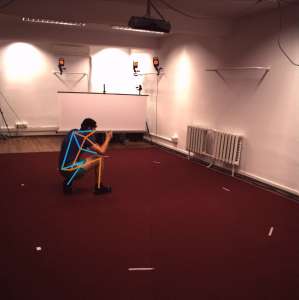}\hspace{\fighspace} & \includegraphics[width=\figsize\textwidth]{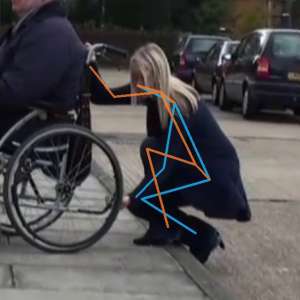}\hspace{\fighspacer}  & \includegraphics[width=\figsize\textwidth]{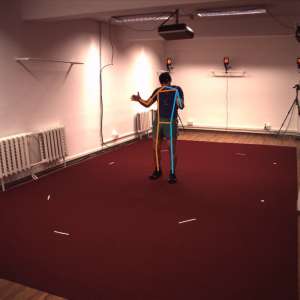}\hspace{\fighspace} & \includegraphics[width=\figsize\textwidth]{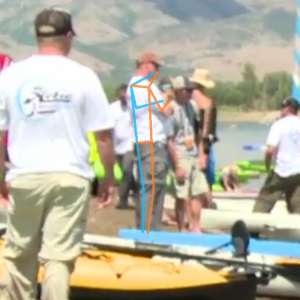}\hspace{\fighspacer} &
\includegraphics[width=\figsize\textwidth]{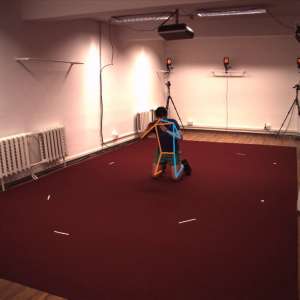}\hspace{\fighspace} & \includegraphics[width=\figsize\textwidth]{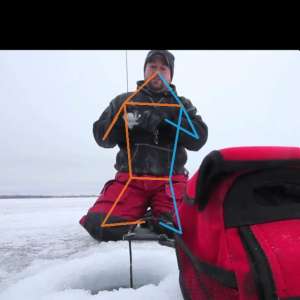}\hspace{\fighspacer} \\

\scriptsize{$C=0.735$}\hspace{\fighspace} & \hspace{\fighspacer} & \scriptsize{$C=0.976$}\hspace{\fighspace} & \hspace{\fighspacer} & \scriptsize{$C=0.448$}\hspace{\fighspace} & \hspace{\fighspacer} \\
\includegraphics[width=\figsize\textwidth]{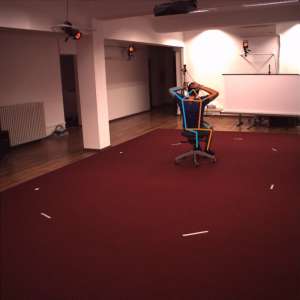}\hspace{\fighspace} & \includegraphics[width=\figsize\textwidth]{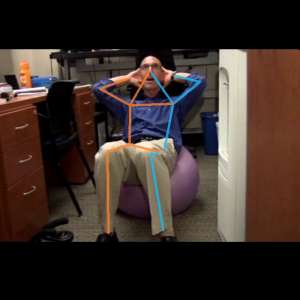}\hspace{\fighspacer}  & \includegraphics[width=\figsize\textwidth]{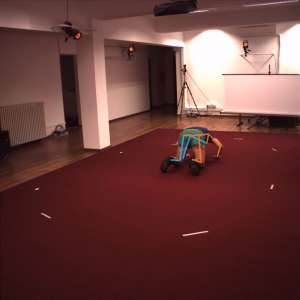}\hspace{\fighspace} & \includegraphics[width=\figsize\textwidth]{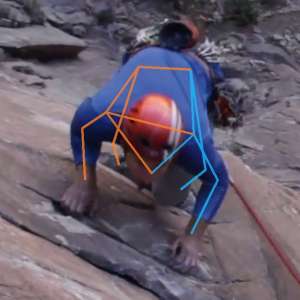}\hspace{\fighspacer} &
\includegraphics[width=\figsize\textwidth]{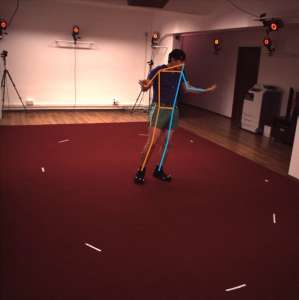}\hspace{\fighspace} & \includegraphics[width=\figsize\textwidth]{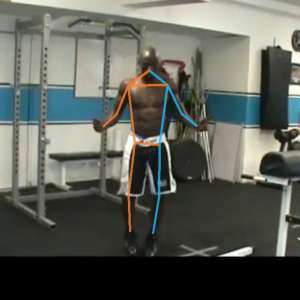}\hspace{\fighspacer} \\

\end{tabular}
\caption{Visualization of pose retrieval results. On each row, we show the query pose on the left for each image pair and the top-$1$ retrieval using the Pr-VIPE model with camera augmentation on the right. We display the retrieval confidences (``C'') and top-$1$ NP-MPJPEs (``E'', if 3D pose groundtruth is available).}
\label{fig:supp_retrieval}
\vspace{-0.2cm}
\end{figure*}

On H3.6M, retrieval confidence is generally high and retrievals are visually accurate. NP-MPJPE is in general smaller on H3.6M compared to 3DHP, since 3DHP has more diverse poses and camera views. The model works reasonably well on 3DHP despite additional variations on pose, viewpoints and subjects. For the pairs R4C3 and R5C3, the subjects are occluded by the chair and the pose inferred by the 2D keypoint detector may not be accurate. Our model is dependent on the result of the 2D keypoint detector. Interestingly, R3C2 and R4C3 show retrievals with large rolls, which is unseen during training. The results on 3DHP demonstrate the generalization capability of our model to unseen poses and views. To test on in-the-wild images, we use the hold-out set of H3.6M to retrieve from 2DHP. The retrieval results demonstrate that Pr-VIPE embeddings can retrieve visually accurate poses from detected 2D keypoints. R7C2 is particularly interesting, as the retrieval has a large change in viewpoint. For the low confidence pairs R6C2 and R7C3, we can see that the arms of the subjects seems to be bent slightly differently. In contrast, the higher confidence retrieval pairs looks visually similar. The results suggest that performance of existing 2D keypoint detectors, such as~\cite{papandreou2018personlab}, is sufficient to train pose embedding models to achieve the view-invariant property in diverse images.

\section{Qualitative Video Alignment Results}\label{sec:video_alignment_qres}

We show that Pr-VIPE can be applied to synchronize action videos from different views from the Penn Action dataset (test set). The videos are synchronized to the pace of a target video (placed in the center of each video array). This allows us to play different videos of the same action at the same pace. The results for different aligned actions are located at \url{https://drive.google.com/open?id=1kTc_UT0Eq0H2ZBgfEoh8qEJMFBouC-Wv}. The alignment procedure for Pr-VIPE is described in Section 4.3.2 in the main paper.


\clearpage
%
%
\bibliographystyle{splncs04}
\bibliography{egbib}

\end{document}